%% file: paper.tex
\documentclass{article}

\PassOptionsToPackage{numbers, compress}{natbib}

\usepackage[preprint]{neurips_2026}

\usepackage[utf8]{inputenc}
\usepackage[T1]{fontenc}
\usepackage{hyperref}
\usepackage{url}
\usepackage{booktabs}
\usepackage{amsfonts}
\usepackage{nicefrac}
\usepackage{microtype}
\usepackage{xcolor}

\usepackage{graphicx}
\usepackage{enumitem}
\usepackage{multirow}
\usepackage{wrapfig}
\usepackage[normalem]{ulem}
\useunder{\uline}{\ul}{}
\usepackage{tabularx}
\usepackage{amsmath}
\usepackage[capitalize,noabbrev]{cleveref}

\let\citet\citep

\title{SR-Prominence: A~Crowdsourced Protocol and Dataset Suite for~Perceptually-Weighted Super-Resolution Artifact Evaluation}

\author{%
  Ivan Molodetskikh\thanks{Corresponding author: \texttt{ivan.molodetskikh@graphics.cs.msu.ru}}\\
  AI Center, Lomonosov Moscow State University\\
  GSP-1, Leninskie Gory, Moscow, 119991, Russia
  \AND
  Kirill Malyshev\\
  AI Center, Lomonosov Moscow State University
  \And
  Mark Mirgaleev\\
  Lomonosov Moscow State University
  \And
  Nikita Zagainov\\
  Innopolis University\\
  Universitetskaya St, 1, Innopolis, Respublika Tatarstan, 420500, Russia
  \And
  Evgeney Bogatyrev\\
  AI Center, Lomonosov Moscow State University
  \And
  Dmitriy Vatolin\\
  MSU Institute for Artificial Intelligence, Lomonosov Moscow State University
}

\begin{document}

\maketitle

\begin{abstract}
Modern image super-resolution methods generate detailed, visually appealing results, but they often introduce visual artifacts: unnatural patterns and texture distortions that degrade perceived quality.
These defects vary widely in perceptual impact---some are barely noticeable, while others are highly disturbing---yet existing detection methods treat them equally.
We propose artifact \emph{prominence} as an evaluative target, defined as the fraction of viewers who judge a highlighted region to contain a noticeable artifact.
We design a crowdsourced annotation protocol and construct \textbf{SR-Prominence}, a dataset suite containing 3,935 artifact masks from DeSRA, Open~Images, Urban100, and a realistic no-ground-truth Urban100-HR setting, annotated with prominence.
Re-annotating DeSRA reveals that 48.2\% of its in-lab binary artifacts are not noticed by a majority of viewers.
Across the suite, we audit SR artifact detectors, image-quality metrics, and SR methods.
We find that classical full-reference metrics, especially SSIM and DISTS, provide surprisingly strong localized prominence signals, whereas no-reference IQA methods and specialized artifact detectors often fail to generalize across datasets and reference settings.
SR-Prominence is released with an objective scoring protocol that allows new metrics to be benchmarked on our suite without further crowdsourcing.
Together, the data and protocols enable SR artifact evaluation to move from binary defect presence toward perceptual impact.
SR-Prominence is available at \url{https://huggingface.co/datasets/imolodetskikh/sr-artifact-prominence}.
\end{abstract}

\begin{figure}[h]
    \centering
    \makebox[\linewidth][c]{%
        \includegraphics[width=0.45\linewidth]{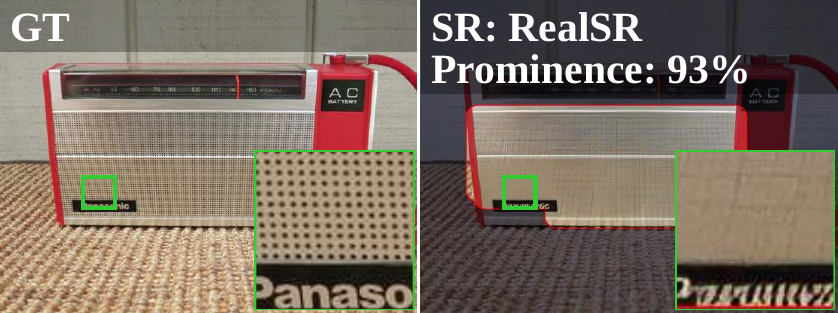}\hspace{0.01\linewidth}%
        \includegraphics[width=0.45\linewidth]{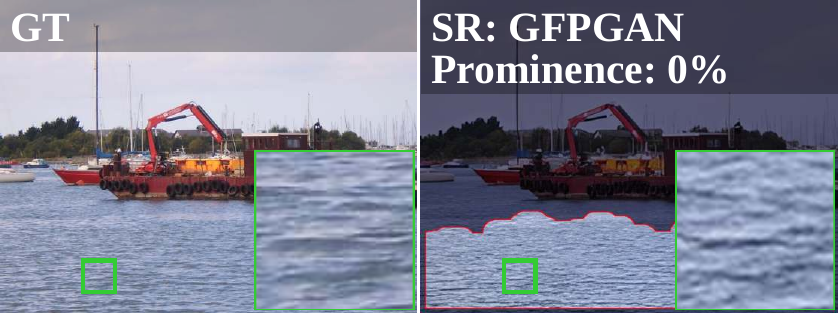}}
    \par\vspace{0.002\linewidth}
    \makebox[\linewidth][c]{%
        \includegraphics[width=0.45\linewidth]{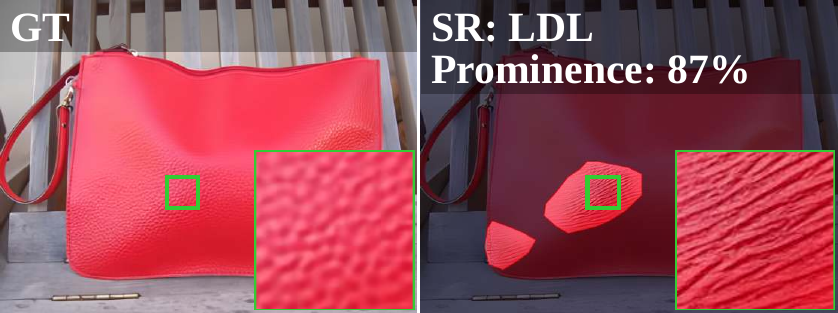}\hspace{0.01\linewidth}%
        \includegraphics[width=0.45\linewidth]{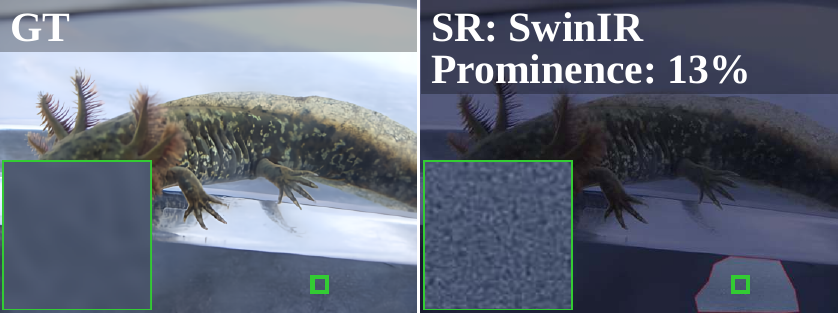}}
    \caption{SR-Prominence artifact examples.
    Rows show Open~Images (top) and DeSRA (bottom) subsets.
    Left: prominent artifacts; RealSR blurred out holes on the radio panel, and LDL reconstructed an incorrect linear pattern on the bag.
    Right: non-prominent artifacts; GFPGAN incorrectly restored a natural water surface, and SwinIR generated a dotted texture artifact on a non-salient floor region.}
    \label{fig:prominence-examples}
\end{figure}

\section{Introduction}
\label{sec:introduction}
Single-image super-resolution (SISR) aims to reconstruct high-resolution (HR) images from low-resolution (LR) inputs.
While modern SR methods have greatly improved perceptual quality, they introduce a critical challenge: visually unpleasant artifacts.
These artifacts---usually unnatural patterns, smeared faces, and texture distortions---degrade perceived quality and hinder adoption.
Even the latest, most capable methods~\citep{yu2024scaling,wang2024exploiting} remain prone to generating artifacts.

Despite SISR's growing popularity, research on detecting SR artifacts remains scarce.
LDL~\citep{liang2022details} and DeSRA~\citep{xie2023desra} identify artifact-prone regions using residual statistics, while segmentation-style approaches such as PAL4VST~\citep{Zhang_2023_ICCV} predict artifact masks from the output image.
These masks localize artifacts, but assign the same label to barely visible texture errors and obvious structural failures.

We use artifact \emph{prominence} for this missing perceptual variable: the fraction of viewers who judge an artifact candidate to contain a noticeable artifact.
Distortions to regular structures such as buildings, or to recognizable objects such as human faces, easily draw attention and can be distressing to viewers, while artifacts on water, grass, and other organic matter can be almost unnoticeable~(\Cref{fig:prominence-examples}).
Treating these different cases as equal carries the risk of overfitting a detection method to low-impact defects while missing the ones that viewers actually notice.

To measure prominence, we design a crowdsourced annotation protocol and use it to construct \textbf{SR\nobreakdash-Prominence}, a four-component dataset suite.
SR-Prominence contains 3,935~artifact masks generated by 15~widely-used SR methods and their variants, each with crowdsourced prominence annotations.
It includes 593~existing masks from the DeSRA dataset~\citep{xie2023desra} and new masks on Open~Images, the standard Urban100 benchmark, and the realistic high-resolution Urban100-HR setting.

Using this suite, we audit artifact detectors, metrics, and SR models.
Our analysis shows that binary labels are not enough for SR artifact evaluation: 48.2\% of DeSRA's in-lab binary artifacts are not noticed by a majority of viewers.
We find that full-reference metrics such as SSIM and DISTS provide strong localized prominence signals, whereas no-reference IQA methods and specialized artifact detectors often fail to generalize.
We also provide a scoring protocol for evaluating new methods on SR-Prominence without further crowdsourcing, a lightweight reference baseline, and a pseudo-GT procedure for applying full-reference metrics when high-resolution ground truth is unavailable.

Our main contributions are the following:
\begin{enumerate}
    \item We introduce artifact \emph{prominence} as a graded target for SR artifact evaluation and design a crowdsourced annotation protocol for collecting prominence labels.
    The protocol includes mask preprocessing for visual assessment and annotation quality control; we analyze response variability to justify the number of assessors.

    \item We construct \textbf{SR-Prominence}, a four-component dataset suite with 3,935~prominence-annotated artifact masks generated by 15~widely-used SR methods with variants.
    The suite covers existing DeSRA masks, diverse natural images from Open~Images, structured scenes from Urban100, and a realistic high-resolution Urban100-HR setting.

    \item We use SR-Prominence to audit SR models for their proneness to generating artifacts, and artifact detection methods, including image-quality metrics.
    We additionally provide a scoring protocol that requires no further crowdsourcing, a pseudo-GT procedure for applying full-reference metrics without HR ground truth, and a small artifact-detection baseline.
\end{enumerate}

\section{Related work}
\label{sec:related}

\textbf{SR artifacts and artifact localization.}
Modern SR methods improve perceptual sharpness but can introduce visual artifacts such as hallucinated structures and texture distortions, especially with adversarial losses or large generative priors~\citep{yu2024scaling,wang2024exploiting,ledig2017photo,wang2021real}.
Detection and mitigation of SISR artifacts has garnered increasing attention because these artifacts reduce perceptual quality.
LDL~\citep{liang2022details} predicts pixel-level artifact maps from local residual statistics.
Xie~et~al.~\citet{xie2023desra} introduced a dataset with SR artifact masks annotated in-lab and proposed DeSRA, which contrasts GAN-SR and MSE-SR outputs to identify and suppress artifact-prone regions.

A complementary line of work treats artifact detection as segmentation, training networks on datasets with pixel-level defect maps.
Given only an input image, these models predict an artifact mask.
Approaches such as PAL4Inpaint~\citep{zhang2022perceptual} and PAL4VST~\citep{Zhang_2023_ICCV} show strong generalization across generative vision tasks by localizing perceptual artifacts.
Concurrently, Ren~et~al.~\citet{ren2025hallucinationscore} propose Hallucination Score that uses a multimodal LLM to provide an image-level hallucination rating for SR outputs, showing strong alignment with human judgments.
The main drawback of this approach is that it lacks spatial localization, which is critical for downstream tasks such as artifact mitigation, SR model fine-tuning, and for handling cases where different regions of an image exhibit different types of artifacts.

Prior SR-artifact work therefore either provides localized but binary masks, or viewer-aligned scores without localized masks.
Our work targets the missing combination: localized SR artifact candidates with graded viewer noticeability.

\textbf{Perceptual image-quality evaluation and human protocols.}
SISR evaluation has traditionally relied on full-reference metrics such as PSNR and SSIM~\citep{wang2004image}, which assess reconstruction fidelity but correlate poorly with perceptual quality---especially for GAN-based outputs where details and artifacts are entangled.
Perceptual metrics such as LPIPS~\citep{zhang2018unreasonable} and DISTS~\citep{ding2022dists} better align with human judgements and are now widely adopted in SR benchmarks.
Some techniques aim to make metrics more artifact-resistant: ERQA~\citep{kirillova2022erqa} evaluates detail restoration by matching edges in reference and test images.
However, in practice existing metrics still fall well short of matching human perception~\cite{borisov2025srmetricsbench}.

Human-judgment datasets and protocols, including PIPAL~\citep{jinjin2020pipal}, LPIPS/BAPPS~\citep{zhang2018unreasonable}, KonIQ-10k~\citep{hosu2020koniq}, and RichHF~\citep{liang2024rich}, show the importance of collecting perceptual labels directly from viewers.
These datasets primarily target global image quality, pairwise perceptual similarity, or generative-image feedback rather than SR artifacts.
SR-Prominence applies the same general principle---human perception should define the evaluation target---to localized artifact mask candidates.

\section{Prominence: definition and annotation protocol}
\label{sec:prominence-protocol}

Existing datasets such as DeSRA~\citep{xie2023desra} contain only binary artifact masks, without information on how noticeable the artifacts are to viewers.
Two masks with similar size can differ sharply in perceptual impact: distorted text or a window grid may be obvious, unlike comparable errors on plants and water.
We use artifact \emph{prominence} for this missing perceptual variable: the fraction of viewers who identify the selected region as containing a noticeable SR artifact.
Each artifact sample in our datasets has a binary mask and a corresponding prominence value obtained via crowdsourced annotation.

\begin{figure}[t]
    \centering
    \includegraphics[width=0.8\linewidth]{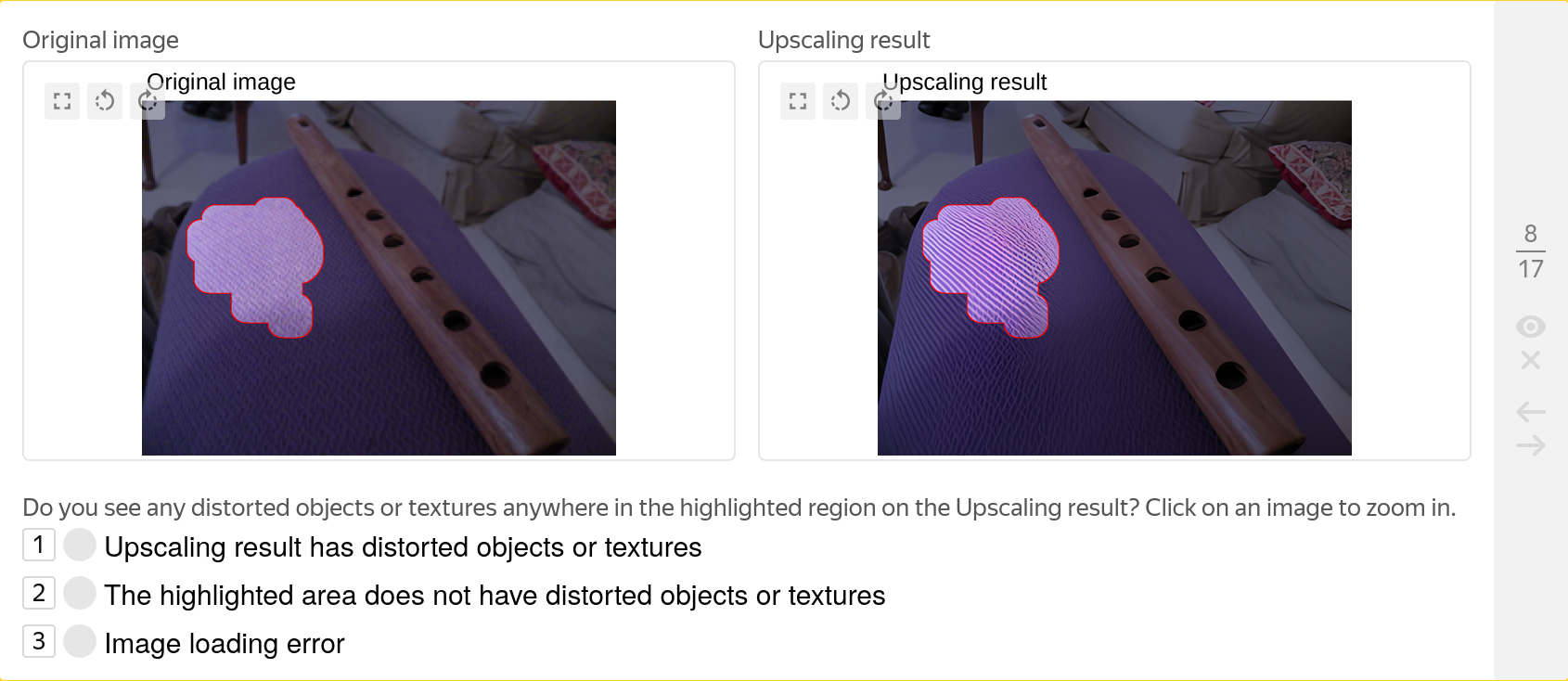}
    \caption{Viewer interface for subjective data collection.}
    \label{fig:subjective-question}
\end{figure}

\subsection{Crowdsourced annotation setup}
\label{sec:annotation-setup}

We used \href{https://tasks.yandex.com}{Yandex.Tasks} to crowdsource the data collection.
Participants view pairs of images labeled ``Original'' and ``Upscaled,'' with the artifact region visually highlighted.
We ask them whether the highlighted region contains a distorted object or texture.
\Cref{fig:subjective-question} shows an example question.

Every image is assessed by 30 different participants.
We compute prominence as the proportion of votes indicating the artifact is present.
Before receiving access to the main questions, participants must answer four training questions, for which the correct answers are explained, followed by four test questions with hidden correct answers.
Afterward, to ensure integrity, every group of 20 questions contains 4 random control questions.
All responses from participants who make mistakes in two or more control questions within an assignment are discarded.

We selected the assessor count using a bootstrap dispersion analysis on 11 images annotated by 250 participants each; details are in~\Cref{sec:appendix-dispersion}.
Few assessors produce unstable prominence estimates; 100 assessors reduce the confidence interval to about \ensuremath{\pm}10\%.
We chose 30 assessors as a practical compromise, giving about \ensuremath{\pm}20\% variability at substantially lower cost.

\subsection{Choice of the ``Original'' image}

In practical SR applications, a full-resolution reference image is unavailable: only the low-resolution input and the SR result are given.
Therefore, assessment cannot rely on exact agreement with an HR ground truth, but should instead focus on the plausibility of restored details and the absence of artifacts.
For the same low-resolution input, many high-resolution outputs may be acceptable.
This is reflected empirically in the mismatch between rankings in SR benchmarks focused on reconstructing the original image and benchmarks focused on perceptual output quality~\citep{kirillova2022erqa,borisov2025srmetricsbench}.

For this reason, our annotation interface uses the low-resolution input, not the high-resolution reference, as the ``Original'' image.
In our protocol, the low-resolution image is upscaled to the target size with nearest-neighbor interpolation before being shown to participants.
Nearest-neighbor interpolation makes it obvious to assessors that they are viewing the low-resolution input.

\subsection{Mask preprocessing}
\label{sec:preprocessing}

\begin{wrapfigure}[10]{r}{0.45\linewidth}
    \vspace{-\baselineskip}
    \centering
    \includegraphics[width=\linewidth]{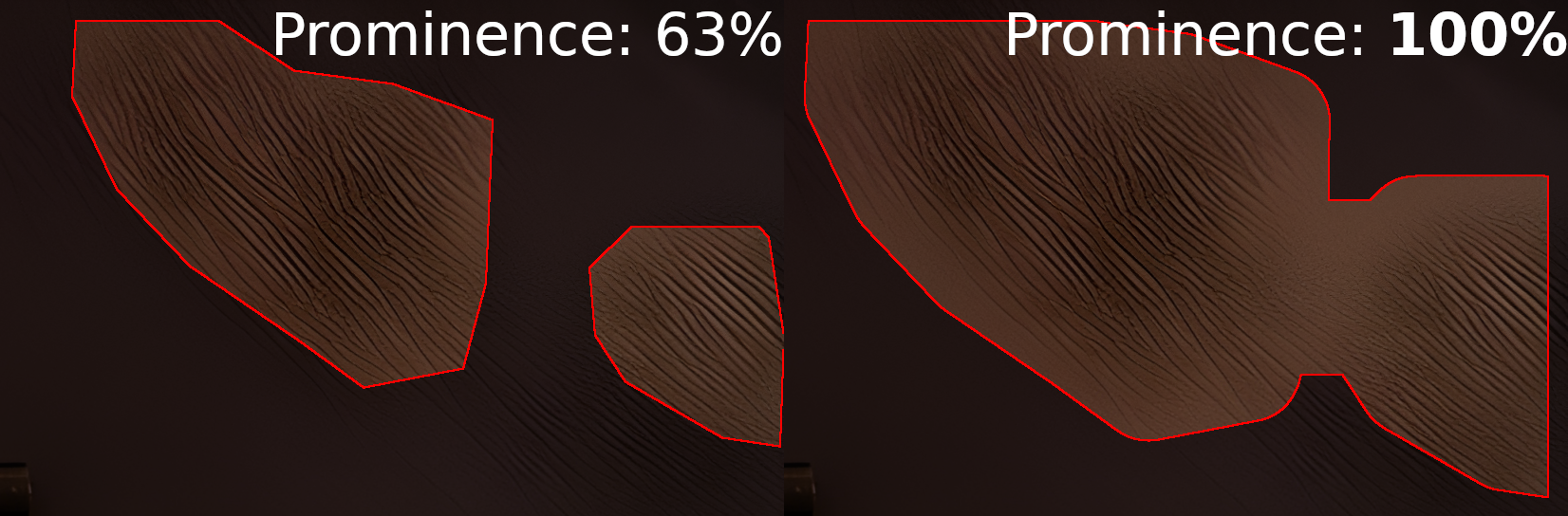}
    \caption{Example of mask preprocessing for human visual assessment.}
    \label{fig:post-example}
\end{wrapfigure}
An artifact-detection method should output a tight mask around an artifact, since such masks are more useful for subsequent analysis and downstream tasks such as automatic correction.
However, tight masks make it harder to visually judge whether the masked area contains an artifact.
Additionally, the raw output from some methods is sparse, making it extra challenging to highlight.
To make masks more suitable for human viewing, we apply morphological operations before showing them to participants:
\begin{enumerate}[nosep]
    \item Open with a 25×25 square kernel to remove small dots in the mask.
    \item Dilate with a 64×64 circular kernel so the mask includes context around an artifact.
    \item Close with a 25×25 square kernel to eliminate holes and step away from the image borders.
\end{enumerate}
The example in \Cref{fig:post-example} shows how a tight mask can make an artifact harder to assess compared with our preprocessing result.
In~\Cref{sec:postprocessing-desra-check} we verify that the effect on already-good masks is negligible.

\subsection{Cropping for large images}

For Prominence-Urban100-HR we run 4× SR directly on the original Urban100 images, so the results extend beyond 4000 pixels wide and 2500 pixels tall.
Our crowdsourcing platform statistics show that most participants use 1920×1080 screens.
Images in full resolution would therefore be downsampled by the browser and viewed at inconsistent zoom.
To remedy this, for Urban100-HR, we crop each image to a padded bounding box around the artifact-mask region before display.
For the majority of the samples, this cropping procedure makes the highlighted region visible at the native scale.
\Cref{sec:u100-hr-cropping} gives the crop parameters and an example.

\section{SR-Prominence dataset suite}
\label{sec:dataset}

\begin{table*}[t]
\centering
\caption{Overview of the SR-Prominence dataset suite.}
\label{tab:sr-prominence-suite}
\resizebox{\textwidth}{!}{%
\begin{tabular}{@{}lrrrrrrr@{}}
\toprule
\textbf{Dataset} & \textbf{\shortstack{Source\\images}} & \textbf{\shortstack{SR\\variants}} & \textbf{\shortstack{Artifact\\masks}} & \textbf{\shortstack{Total\\workers}} & \textbf{\shortstack{Valid\\workers}} & \textbf{\shortstack{Mean\\prominence}} & \textbf{\shortstack{Prominent\\masks}} \\
\midrule
Prominence-DeSRA & 528 & 3 & 593 & 276 & 176 & 49.0\% & 307 (51.8\%) \\
Prominence-OpenImages & 547 & 15 & 1523 & 261 & 183 & 37.1\% & 535 (35.1\%) \\
Prominence-Urban100 & 68 & 19 & 873 & 191 & 110 & 35.5\% & 241 (27.6\%) \\
Prominence-Urban100-HR & 94 & 16 & 946 & 164 & 94 & 24.7\% & 177 (18.7\%) \\
\midrule
Combined & 1237 & 19 & 3935 & 662 & 425 & 35.6\% & 1260 (32.0\%) \\
\bottomrule
\end{tabular}%
}
\end{table*}

We apply the protocol described above to annotate a dataset suite of four complementary components.
Each dataset component contains low-resolution input images, high-resolution SR upscaling results, binary artifact masks, and crowdsourced prominence values for every artifact mask.
\Cref{sec:dataset-examples} shows example masks from each component across the prominence range.
Together the suite contains 3,935 masks from 15 widely-used SR methods.
The Urban100 components additionally evaluate two real-time SR models (SPAN and RLFN) that serve as pseudo-GT references in~\Cref{sec:pseudo-gt}, plus a SUPIR half-precision variant and the full SeeSR model.
The dataset is publicly available at \url{https://huggingface.co/datasets/imolodetskikh/sr-artifact-prominence}.

\Cref{tab:sr-prominence-suite} summarizes the components.
The worker columns report unique crowd workers who participated and who remained after quality-control filtering; in the combined row, workers shared across components are counted once.
The final column reports masks with prominence $\ge$ 50\%, i.e. at least half of valid workers confirmed the presence of an artifact.

Together, these components separate four evaluation questions: whether existing binary artifact datasets are viewer-noticeable, how SRs and detectors behave on diverse natural images, how they handle structured urban content, and how the results change in a realistic high-resolution setting.

\subsection{Candidate-mask collection for scalable annotation}
\label{sec:mask-collection}

After obtaining the Prominence-DeSRA results, it became clear that a more extensive artifact dataset is necessary, covering more image content types and a wide selection of contemporary SR methods.
The other three parts of our dataset suite use an automatic mask collection procedure that we designed to be able to scale to many source images and SR models.
Given SR upscaling results and reference images, we run existing artifact detection and image-quality assessment methods to obtain heatmaps.
The heatmaps are thresholded with fixed DeSRA-calibrated thresholds described in~\Cref{sec:uncurated-benchmark}, producing candidate artifact masks.
We measure the mean heatmap value inside each mask and pick the 10 strongest artifacts per metric per SR.
The masks are then preprocessed following~\Cref{sec:preprocessing} and proceed to crowdsource annotation.

This fully automatic procedure scales to a large number of source images and SR methods.
No expensive manual human mask drawing is required, and manual selection bias is reduced.

\subsection{Prominence-DeSRA}

For Prominence-DeSRA, we collected prominence annotations for all 593 artifact masks from the DeSRA~\citep{xie2023desra} dataset.
It lets us directly test whether binary in-lab artifact masks correspond to viewer-noticeable artifacts, but it is not sufficient as a general prominence benchmark.
It covers only three SR methods and does not test modern diffusion- or transformer-based SR systems across diverse image content.
We therefore extend the suite along three axes that are important for prominence-aware SR evaluation: broader natural-image diversity, structured scenes where localized artifacts are especially visible, and a deployment-style setting without HR ground truth.

\subsection{Prominence-OpenImages}

Prominence-OpenImages targets diverse natural-image SR artifacts.
We randomly selected 2,101 source photos from Open Images~\citep{kuznetsova2020open}, each at 1024×768 pixels, downsampled them by 4× with bicubic interpolation, and upsampled them with 15 popular SR methods and variants.

The first 697 masks came from manual annotation and from existing visual-quality metrics or artifact detectors, including SSIM, DISTS, LPIPS, LDL, and DeSRA.
Afterward, we used our fully-automatic mask proposal pipeline with no manual curation to collect the remaining 826 masks.
Of the 2,101 source photos, 547 contributed at least one selected mask to the Prominence-OpenImages set.
The summary~\Cref{tab:summary-sr-combined,tab:summary-met-combined} report SR and metric results only on the uncurated subset to avoid bias.

\subsection{Prominence-Urban100}

Prominence-OpenImages provides broad natural-image coverage, but many prominent SR artifacts occur in structured content such as text, façades, window grids, signs, and repeated patterns.
Prominence-Urban100 targets this setting using a standard SR benchmark whose regular structures are known to stress SR methods.
It lets the suite test whether prominence depends on semantic and geometric context rather than only on generic image diversity.
This component is based on the standard Urban100 SR test set~\citep{Huang_2015_CVPR}.
It contains 873 masks from 68 source images and 19 SR settings, collected with the fully uncurated procedure.

\subsection{Prominence-Urban100-HR}
\label{sec:prominence-urban100-hr}

Prominence-Urban100-HR is a more challenging and realistic variant of Prominence-Urban100.
Instead of the downsample-then-upsample setting, we took the original high-resolution Urban100 images and upsampled them as-is, with no additional processing or synthetic degradation.
This part of the dataset contains 946 masks from 94 source images and 16 SR settings (fewer, as the VRAM constraints allowed).
Its mean prominence is lower than the other dataset components, which is expected as the 4× larger source images make details clearer and easier to upscale.

No higher-resolution ground truth exists for these outputs, reducing SR-overfitting risk.
The human annotations do not require such a reference, since workers compare the SR output to the LR-derived original as in \Cref{sec:prominence-protocol}.
For full-reference metric evaluation only, we use the RLFN pseudo-GT protocol described in \Cref{sec:pseudo-gt}.

\section{Benchmark tasks and scoring}

We will now define SR-Prominence benchmark tasks, scoring, and reference protocols.

\subsection{Uncurated top-artifact benchmark}
\label{sec:uncurated-benchmark}

The uncurated benchmark evaluates whether SR models and detection methods expose artifacts that viewers actually notice.
For each dataset component, we report two complementary views.
Detector tables group masks by the metric that found them and measure whether that metric finds prominent artifacts.
SR tables group masks by the SR model that produced the output and measure whether an SR model tends to produce prominent artifacts.
For detector tables, higher mean prominence and more confident masks are better.
For SR tables, lower values are better.

This evaluation required thresholding raw metric output to get candidate masks as described in~\Cref{sec:mask-collection}.
Following~\citet{xie2023desra}, we selected the thresholds for each method by maximizing the Precision~×~Recall product on the prominent subset of the DeSRA dataset (307 masks).

To account for both mean prominence and the number of confident masks, \Cref{tab:summary-met-combined} additionally reports their product in the ``Prom.~×~Conf.'' column.
This combined score rewards detectors that find many prominent artifacts rather than only a few high-prominence examples.
For example, we tested LDL at two lower thresholds, which increased the total masks found, but they mainly captured non-prominent artifacts, yielding even worse combined score.
This follows the same intuition as the Precision~×~Recall threshold criterion: both coverage and confidence matter.

For per-SR tables, when multiple metrics found an artifact mask on the same SR image, we deduplicated these masks by selecting the one with the highest prominence value.
This way, SR outputs containing prominent artifacts are not overcounted in the scores.

\subsection{Threshold-free scoring approach}
\label{sec:score}

\begin{table}[t]
    \centering
    \caption{Crowd-sourced prominence results across SR models (DeSRA).}
    \label{tab:summary-sr-desra}
    \resizebox{.7\linewidth}{!}{%
            \begin{tabular}{@{}llrrr@{}}
            \toprule
            \textbf{SR} & \textbf{Family} & \textbf{Masks} & \textbf{Mean Prominence}$^\downarrow$ & \textbf{Conf. Masks}$^\downarrow$ \\
            \midrule
            SwinIR~\citep{liang2021swinir} & Transformer & 206 & \textbf{42.15\%} & \textbf{78} \\
            RealESRGAN~\citep{wang2021real} & CNN & 199 & \underline{46.52\%} & \underline{97} \\
            LDL-SR~\cite{liang2022details} & CNN & 188 & 59.28\% & 132 \\
            \bottomrule
            \end{tabular}%
    }
\end{table}

The uncurated benchmark above evaluates thresholded candidate masks that were shown to crowd workers.
It is the primary benchmark because the final target is viewer prominence.
However, it is expensive to extend to every new SR or detection method, since each additional mask requires crowdsourced annotations.
We therefore propose a threshold-free objective score that measures how well an artifact detection method evaluates artifact prominence on our dataset suite.

A detection method is run on all input images from the dataset, producing spatial heatmaps for each of them.
For each annotated mask, we compute the median heatmap value inside the mask and subtract the median heatmap value outside the mask.
Subtracting the outside value makes the score less sensitive to cross-image and cross-SR variation.
For masks that were dilated during the visual-assessment preprocessing in~\Cref{sec:preprocessing}, we first erode them back, so that the inside region better matches the original artifact localization.

We then compute Spearman correlation between this contrast value and the crowdsourced prominence over all masks in a dataset component:
\(
    \mathrm{SRCC}(\text{inside}_{p50}-\text{outside}_{p50}, \text{GT prominence}).
\)
A high positive correlation means that the detector not only responds inside the annotated regions, but responds more strongly for artifacts noticed by more viewers.
For binary detectors, we apply the same calculation to their binary masks.

This score is designed for rapid benchmarking after the prominence annotations have been collected.
It cannot reveal artifacts that no detector proposed.
Instead, it provides a way to compare heatmaps and reference choices without additional crowdsourcing.
\Cref{tab:heatmap-prominence-contrast-reference} shows this score for the evaluated methods.
For verification, we compared scores with the Prom.~×~Conf. crowdsourced rankings on Open~Images, Urban100, and Urban100-HR, obtaining Spearman correlations 0.886, 0.750, 0.786.

\subsection{Adapting full-reference metrics to no-HR settings with pseudo-GT}
\label{sec:pseudo-gt}

Full-reference metrics provide more-accurate detail restoration quality scores for SR by using pixel-level information from the reference image.
The use of such metrics in SR creates difficulties, however, since the SR-output resolution is higher than that of the original low-resolution frame.
This restriction is unavoidable for Prominence-Urban100-HR, which has no high-resolution ground truth.
Existing reduced-reference SR quality metrics address this mismatch by upscaling the low-resolution input to the SR resolution, for example, with bicubic~\citep{zhou2023structure} or bilinear~\citep{lin2024pfiqa} interpolation.

To employ full-reference metrics in this setting, we use the following pipeline.
We apply a lightweight SR method to the original low-resolution frame, thereby obtaining a pseudo-GT, and then calculate the metric between this pseudo-GT and the SR output.
The pseudo-GT model should be conservative: it may trail heavier SR models in visual quality, but it should avoid producing prominent artifacts of its own.
Real-time SR methods such as SPAN~\citep{wan2024swift} and RLFN~\citep{Fangyuan2022RLFN} are natural candidates for this role.
In our experiments we use RLFN, which is less artifact-prone than SPAN on Urban100 in \Cref{tab:summary-sr-combined}.

When serving as pseudo-GT for full-reference metrics, the artifact-detection performance drop is small compared with using the original HR frames.
We characterize this approximation in \Cref{tab:heatmap-prominence-contrast-reference}.

\subsection{Baseline artifact detection method}
\label{sec:artifact-prominence-metric}

To accompany the dataset and evaluation protocol, we provide a simple reference baseline that predicts a spatial artifact-prominence heatmap for a super-resolved image.
The baseline computes three complementary heatmaps: block-wise DISTS, which is strong in our metric analysis, and two features adapted from JPEG~AI artifact work~\citet{Tsereh2024JPEGAI}: \textit{ssm\_jup}, an RGB adaptation of a local residual-variance detector, and \textit{bd\_jup}, a block-wise combination of LPIPS and ERQA.
A shallow MLP fuses these features independently at each pixel and is trained on 374 Prominence-OpenImages artifact samples to match the crowdsourced prominence inside the annotated mask and zero outside it.

Despite this limited training setup, the baseline generalizes to held-out OpenImages examples and to the Urban100, Urban100-HR, and DeSRA components, achieving the best average rank in the threshold-free prominence score in~\Cref{tab:heatmap-prominence-contrast-reference}.
We describe our baseline in detail and provide ablation experiments in~\Cref{sec:baseline-details}.

\section{Empirical findings from prominence-aware evaluation}
\label{sec:experiments}

We use SR-Prominence to audit binary masks, detectors, image-quality metrics, and SR models.

\subsection{Binary masks are insufficient annotation for SR artifacts}

\Cref{tab:summary-sr-desra} shows per-SR prominence results that we obtained on the DeSRA dataset.
This dataset provides artifact masks for three SR methods annotated in-lab, so it was surprising to learn that only in 307 of 593 masks at least half of workers confirmed the artifact under our protocol.
Equivalently, 48.2\% of DeSRA binary artifacts are not noticed by a majority of crowd workers.
This result is the cleanest evidence in our suite that binary artifact masks are insufficient for SR artifact assessment.

\subsection{Structural and perceptual similarity metrics outperform specialized detectors}
\label{sec:findings-metrics}

\begin{table*}[t]
\centering
\renewcommand{\arraystretch}{0.9}
\caption{Crowd-sourced prominence results across artifact detection methods.}
\label{tab:summary-met-combined}
\resizebox{.99\textwidth}{!}{%
\begin{tabular}{@{}l|rrrr|rrrr|rrrr|r@{}}
\toprule
 & \multicolumn{4}{c|}{\textbf{Open Images}} & \multicolumn{4}{c|}{\textbf{Urban100}} & \multicolumn{4}{c|}{\textbf{Urban100-HR}} &  \\ \cmidrule(lr){2-5} \cmidrule(lr){6-9} \cmidrule(lr){10-13}
\vtop{\hbox{\strut}\hbox{\strut}\hbox{\strut \textbf{Method}}} & \vtop{\hbox{\strut}\hbox{\strut Masks}\hbox{\strut Found}} & \vtop{\setbox0\hbox{\strut Prominence$^\uparrow$}\hbox{\strut}\hbox to\wd0{\hss\strut Mean\hss}\copy0} & \vtop{\hbox{\strut Conf.}\hbox{\strut Masks}\hbox{\strut Found$^\uparrow$}} & \vtop{\hbox{\strut}\hbox{\strut Prom.~$\times$}\hbox{\strut Conf.$^\uparrow$}} & \vtop{\hbox{\strut}\hbox{\strut Masks}\hbox{\strut Found}} & \vtop{\setbox0\hbox{\strut Prominence$^\uparrow$}\hbox{\strut}\hbox to\wd0{\hss\strut Mean\hss}\copy0} & \vtop{\hbox{\strut Conf.}\hbox{\strut Masks}\hbox{\strut Found$^\uparrow$}} & \vtop{\hbox{\strut}\hbox{\strut Prom.~$\times$}\hbox{\strut Conf.$^\uparrow$}} & \vtop{\hbox{\strut}\hbox{\strut Masks}\hbox{\strut Found}} & \vtop{\setbox0\hbox{\strut Prominence$^\uparrow$}\hbox{\strut}\hbox to\wd0{\hss\strut Mean\hss}\copy0} & \vtop{\hbox{\strut Conf.}\hbox{\strut Masks}\hbox{\strut Found$^\uparrow$}} & \vtop{\hbox{\strut}\hbox{\strut Prom.~$\times$}\hbox{\strut Conf.$^\uparrow$}} & \vtop{\hbox{\strut}\hbox{\strut \textbf{Avg.}}\hbox{\strut \textbf{Rank}$^\downarrow$}} \\ \midrule
bd\_jup {\footnotesize (t=0.1)} & 150 & 22.20\% & 27 & 5.99\% & 190 & 32.28\% & 37 & 11.94\% & 160 & 12.58\% & 6 & 0.75\% & 6.3 \\
LDL {\footnotesize (t=0.005)} & 23 & \textbf{70.23\%} & 17 & 11.94\% & 1 & 6.67\% & 0 & 0.00\% & 113 & 24.32\% & 15 & 3.65\% & 5.7 \\
ssm\_jup {\footnotesize (t=0.15)} & 150 & 26.91\% & 36 & 9.69\% & 155 & 34.03\% & 41 & 13.95\% & 160 & 11.44\% & 8 & 0.92\% & 5.3 \\
DeSRA & 120 & 34.72\% & 38 & 13.19\% & 10 & \textbf{56.83\%} & 6 & 3.41\% & 39 & \textbf{48.35\%} & 20 & 9.67\% & 4.3 \\
\textbf{Baseline} {\footnotesize (t=0.3)} & 139 & \underline{43.07\%} & \underline{55} & \textbf{23.69\%} & 140 & \underline{39.29\%} & 46 & \underline{18.07\%} & 160 & 27.59\% & 32 & 8.83\% & \underline{2.3} \\
DISTS {\footnotesize (t=0.25)} & 148 & 41.31\% & \textbf{57} & \underline{23.55\%} & 190 & 34.28\% & \underline{47} & 16.11\% & 160 & 32.40\% & \textbf{46} & \textbf{14.90\%} & \textbf{2.0} \\
SSIM {\footnotesize (t=0.55)} & 114 & 41.73\% & 47 & 19.61\% & 187 & 37.02\% & \textbf{58} & \textbf{21.47\%} & 154 & \underline{32.77\%} & \underline{41} & \underline{13.44\%} & \textbf{2.0} \\
\bottomrule
\end{tabular}%
}
\end{table*}

\begin{table*}[t]
\centering
\renewcommand{\arraystretch}{0.9}
\caption{Crowd-sourced prominence results across SR models.}
\label{tab:summary-sr-combined}
\resizebox{.99\textwidth}{!}{%
\begin{tabular}{@{}ll|rrr|rrr|rrr@{}}
\toprule
 &  & \multicolumn{3}{c|}{\textbf{Open Images}} & \multicolumn{3}{c|}{\textbf{Urban100}} & \multicolumn{3}{c}{\textbf{Urban100-HR}} \\ \cmidrule(lr){3-5} \cmidrule(lr){6-8} \cmidrule(lr){9-11}
\vtop{\hbox{\strut}\hbox{\strut}\hbox{\strut \textbf{SR}}} & \vtop{\hbox{\strut}\hbox{\strut}\hbox{\strut \textbf{Family}}} & \vtop{\hbox{\strut}\hbox{\strut Masks}\hbox{\strut Found}} & \vtop{\setbox0\hbox{\strut Prominence$^\downarrow$}\hbox{\strut}\hbox to\wd0{\hss\strut Mean\hss}\copy0} & \vtop{\hbox{\strut Conf.}\hbox{\strut Masks}\hbox{\strut Found$^\downarrow$}} & \vtop{\hbox{\strut}\hbox{\strut Masks}\hbox{\strut Found}} & \vtop{\setbox0\hbox{\strut Prominence$^\downarrow$}\hbox{\strut}\hbox to\wd0{\hss\strut Mean\hss}\copy0} & \vtop{\hbox{\strut Conf.}\hbox{\strut Masks}\hbox{\strut Found$^\downarrow$}} & \vtop{\hbox{\strut}\hbox{\strut Masks}\hbox{\strut Found}} & \vtop{\setbox0\hbox{\strut Prominence$^\downarrow$}\hbox{\strut}\hbox to\wd0{\hss\strut Mean\hss}\copy0} & \vtop{\hbox{\strut Conf.}\hbox{\strut Masks}\hbox{\strut Found$^\downarrow$}} \\ \midrule
LDL-SR~\cite{liang2022details} & CNN & 52 & 66.19\% & 19 & 52 & 54.66\% & 17 & 60 & 39.20\% & 9 \\
RealESRGAN~\citep{wang2021real} & CNN & 61 & 48.42\% & 19 & 51 & 47.01\% & 12 & 55 & 31.71\% & 11 \\
SwinIR~\citep{liang2021swinir} & Transformer & 59 & 41.09\% & 17 & 51 & 39.49\% & 10 & 61 & 33.54\% & 12 \\
GFPGAN~\citep{wang2021towards} & CNN & 45 & 32.74\% & 11 & 54 & 52.17\% & 14 & 56 & 39.94\% & 12 \\
SUPIR~\citep{yu2024scaling} & Diffusion & 70 & 45.29\% & 20 & 50 & 26.90\% & 4 & - & - & - \\
StableSR~\citep{wang2024exploiting} & Diffusion & 50 & 33.18\% & 13 & 50 & 35.39\% & 7 & 62 & 26.81\% & 11 \\
OSEDiff~\cite{wu2024one} & Diffusion & 50 & 30.33\% & 12 & 50 & 42.24\% & 11 & 55 & 22.92\% & 7 \\
RealSR~\citep{Ji_2020_RealSR} & CNN & 51 & 28.70\% & 10 & 47 & 48.22\% & 14 & 67 & 27.19\% & 5 \\
PASD-SDXL~\cite{yang2024pixel} & Diffusion & 69 & 35.95\% & 12 & 51 & 27.77\% & 6 & 70 & 29.22\% & 10 \\
SeeSR-turbo~\citep{wu2024seesr} & Diffusion & 61 & 32.34\% & 14 & 50 & 26.82\% & 4 & 66 & 25.47\% & 9 \\
SinSR~\citep{wang2024sinsr} & Diffusion & 60 & 19.39\% & \underline{3} & 46 & 43.87\% & 11 & 55 & 28.26\% & 5 \\
SPAN~\citep{wan2024swift} & CNN & - & - & - & 36 & 43.97\% & 9 & 48 & 21.47\% & \underline{3} \\
StableSR-turbo~\citep{wang2024exploiting} & Diffusion & 60 & 28.55\% & 14 & 50 & 24.33\% & 2 & 56 & 17.01\% & \underline{3} \\
RLFN~\citep{Fangyuan2022RLFN} & CNN & - & - & - & 36 & 31.10\% & 4 & - & - & - \\
SUPIR-half~\citep{yu2024scaling} & Diffusion & - & - & - & 50 & 26.08\% & 4 & - & - & - \\
ResShift~\citep{yue2023resshift} & Diffusion & 60 & 20.22\% & 7 & 42 & 36.99\% & 6 & 59 & 21.05\% & 4 \\
SeeSR~\citep{wu2024seesr} & Diffusion & - & - & - & 50 & 26.57\% & 4 & 61 & 23.42\% & 4 \\
HAT-L~\citep{chen2023hat} & Transformer & 53 & \underline{13.53\%} & \textbf{1} & 29 & \underline{22.80\%} & \underline{1} & 60 & \underline{15.48\%} & \textbf{1} \\
DRCT~\citep{Hsu_2024_CVPR} & Transformer & 43 & \textbf{11.85\%} & \textbf{1} & 28 & \textbf{21.47\%} & \textbf{0} & 55 & \textbf{12.38\%} & \textbf{1} \\
\bottomrule
\end{tabular}%
}
\end{table*}

\begin{table*}[t]
\centering
\renewcommand{\arraystretch}{0.9}
    \caption{Threshold-free prominence score described in~\Cref{sec:score}.
    RLFN columns use the pseudo-GT protocol from~\Cref{sec:pseudo-gt}.}
\label{tab:heatmap-prominence-contrast-reference}
\resizebox{.99\textwidth}{!}{%
\begin{tabular}{@{}l|r r r r r r r |r@{}}
\toprule
\textbf{} & \multicolumn{2}{c}{\textbf{Open Images}} & \multicolumn{2}{c}{\textbf{Urban100}} & \multicolumn{1}{c}{\textbf{Urban100-HR}} & \multicolumn{2}{c}{\textbf{DeSRA}} & \multicolumn{1}{|c}{\textbf{Avg.}} \\ \cmidrule{1-8}
\multicolumn{1}{r|}{\textbf{Reference Input}} & \textbf{Original HR} & \textbf{RLFN} & \textbf{Original HR} & \textbf{RLFN} & \textbf{RLFN} & \textbf{MSE-SR} & \textbf{RLFN} & \textbf{Rank}$^\downarrow$ \\ \midrule
TOPIQ \footnotesize{(no-ref)} & -0.096 & -0.096 & -0.099 & -0.099 & -0.015 & -0.114 & -0.114 & 11.9 \\
PaQ-2-PiQ \footnotesize{(no-ref)} & -0.097 & -0.097 & -0.222 & -0.222 & 0.024 & -0.088 & -0.088 & 11.7 \\
ERQA & 0.142 & -0.134 & 0.197 & 0.138 & 0.025 & -0.024 & 0.023 & 8.7 \\
PAL4VST \footnotesize{(bin., no-ref)} & 0.136 & 0.136 & 0.145 & 0.145 & N/A & 0.043 & 0.043 & 7.7 \\
PAL4Inpaint \footnotesize{(bin., no-ref)} & 0.067 & 0.067 & 0.156 & 0.156 & 0.112 & 0.109 & 0.109 & 7.4 \\
DeSRA & 0.303 & 0.299 & 0.019 & 0.049 & 0.340 & 0.021 & -0.003 & 7.3 \\
bd\_jup & 0.346 & 0.177 & 0.236 & 0.245 & 0.267 & -0.105 & -0.098 & 6.6 \\
LPIPS & {\ul 0.446} & 0.304 & 0.213 & 0.234 & \textbf{0.446} & -0.138 & -0.132 & 6.0 \\
ssm\_jup & 0.321 & 0.237 & 0.092 & 0.099 & 0.099 & \textbf{0.342} & \textbf{0.331} & 6.0 \\
LDL & 0.288 & 0.233 & 0.158 & 0.200 & 0.255 & 0.193 & 0.177 & 5.9 \\
SSIM & 0.312 & 0.297 & \textbf{0.305} & {\ul 0.264} & 0.348 & -0.010 & -0.020 & 4.9 \\
DISTS & 0.415 & {\ul 0.370} & 0.236 & \textbf{0.283} & {\ul 0.401} & -0.039 & -0.009 & {\ul 4.1} \\
\textbf{Baseline} & \textbf{0.555} & \textbf{0.374} & {\ul 0.256} & 0.241 & 0.394 & {\ul 0.213} & {\ul 0.212} & \textbf{2.1} \\
\bottomrule
\end{tabular}%
}
\end{table*}

\Cref{tab:summary-met-combined} shows the results of the uncurated crowdsourced detector benchmark.

Across the three uncurated dataset components, DISTS and SSIM are the strongest existing metrics by average rank, showing better performance than DeSRA and LDL---purpose-made SR artifact detectors.
This is notable as neither metric was designed for SR-artifact assessment, especially so SSIM, a classical image-quality assessment metric.
DISTS is a learned perceptual similarity metric trained on natural images to account for texture distortions, which makes its strong performance less surprising.
Together, these results show that full-reference structural and perceptual similarity metrics contain the most useful information about viewer-noticeable SR failures.

The threshold-free analysis in \Cref{tab:heatmap-prominence-contrast-reference} supports the same conclusion from a different viewpoint, and includes a wider selection of metrics.
DISTS and SSIM also remain competitive when the HR reference is replaced with the RLFN pseudo-GT.

\subsection{Existing no-reference methods fail to model artifact prominence}
\label{sec:findings-nr}

No-reference prominence evaluation is an inherently harder task: it requires a good understanding of the image to tell apart artifacts in different semantic contexts.
In \Cref{tab:heatmap-prominence-contrast-reference}, no-reference methods perform worse than the methods that can use an HR or pseudo-GT reference, with NR IQA methods in particular having correlations close to zero or negative on several components.
This is expected for whole-image no-reference IQA metrics, which are hard to adapt to localized prominence evaluation because their original target is global image quality.
The same limitation appears for no-reference artifact detectors PAL4Inpaint and PAL4VST: despite being designed to localize perceptual artifacts, and despite PAL4VST including SR among its target distortion sets, they rank prominence poorly.

\subsection{Artifact-aware SR training does not guarantee low-prominence artifacts}
\label{sec:findings-sr}

\Cref{tab:summary-sr-combined} shows the uncurated crowdsourced benchmark results grouped by the SR model that produced each candidate artifact mask.
Surprisingly, LDL-SR is the weakest method overall, with the highest mean prominence and many confident masks on both Open~Images and Urban100, despite being specifically trained for artifact prevention.
This is consistent with the Prominence-DeSRA result in \Cref{tab:summary-sr-desra}, where LDL-SR also produces the most prominent artifacts.

The most robust methods in our benchmark are DRCT and HAT-L.
They have the lowest mean prominence on every reported component and produce at most one confident mask in each setting, with DRCT producing none on Urban100.
This suggests that strong reconstruction-oriented Transformer SR can avoid prominent artifacts more reliably than generative methods.

The Urban100 and Urban100-HR results separate two difficult structured-scene settings.
In the standard downsample-then-upsample setting, most methods show higher prominence on Urban100 than on Urban100-HR, with the largest drops of 19--22 percentage points for OSEDiff, RealSR, and SPAN.
This suggests that the synthetic benchmark setting can amplify visible failures on structured content.
At the same time, the no-HR Urban100-HR setting is not merely easier: LDL-SR, GFPGAN, and SwinIR still produce many prominent masks, while HAT-L and DRCT remain robust.
The two Urban100 components therefore offer complementary views of SR robustness on structured content.

\subsection{Artifact type and semantic context matter}
\label{sec:artifact-types}

To characterize where prominent SR artifacts occur, we additionally annotated artifact types and semantic context with Qwen 3 VLM~(\Cref{sec:qwen-type-context-summary}).
Among the most prominent types are hallucinated texture and text aberrations, with plastic texture being less noticeable.
By semantic context, artifacts in art and text images are the most prominent, while artifacts in nature images are the least.
These results support the design of SR-Prominence: it combines broad natural-image diversity with urban scenes, covering contexts in which the same nominal artifacts can have different perceptual impact.

\section{Limitations and future work}
\label{sec:limitations}

The main limitation of SR-Prominence is the lack of precise artifact masks, as delineating exact artifact boundaries is ambiguous even for human annotators.
Instead, we seed masks using existing methods, introducing some inaccuracy.
Consequently, models trained on this data may have lower performance due to imperfect supervision.
Seeding also misses artifacts that no seed detectors found.
In~\Cref{sec:seed-detector-coverage} we study the impact on hand-annotated artifacts and find such misses to be rare in our setup.

Our pseudo-GT procedure is also only an approximation.
It allows full-reference metrics to be applied in a realistic no-HR setting, but can produce false positives when the lightweight SR model used as pseudo-GT fails to reconstruct fine textures.
While it is consistent with existing work on reduced-reference quality assessment, future research could investigate using the low-resolution reference as is, without upsampling.

Future work could also extend prominence modeling to video super-resolution.
We consider images independently and do not address temporal artifacts such as flickering.
Another direction is semantic artifacts: higher-capacity models such as SUPIR move from simple texture distortions toward more semantic failures like object replacement, which may need adjustments to annotation and evaluation.

\section{Conclusion}
\label{sec:conclusion}

We introduced artifact prominence as a viewer-centered target for super-resolution artifact evaluation and constructed SR-Prominence, a four-component dataset suite of prominence-annotated artifact masks.
Instead of treating all localized defects as equal, prominence measures how often viewers judge a region to contain a noticeable artifact.
Our results show that binary artifact labels are insufficient: roughly half of DeSRA binary masks are not noticed by a majority of crowd workers.

Across the broader suite, full-reference metrics provide strong localized prominence signals, while no-reference IQA methods and specialized artifact detectors often fail to generalize.
The SR-model audit further shows that artifact-aware training does not necessarily reduce prominent artifacts.

SR-Prominence is intended to make SR artifact evaluation more perceptually grounded.
The released annotations, scoring protocol, pseudo-GT procedure, and reference baseline allow future detectors and SR methods to be evaluated against viewer noticeability without repeating the full crowdsourcing process.
More broadly, our findings suggest that SR artifact evaluation should move beyond binary defect presence and account for which artifacts viewers actually notice.

\section*{Acknowledgments}
The work of Ivan Molodetskikh, Kirill Malyshev and Evgeney Bogatyrev was supported by the Ministry of Economic Development of the Russian Federation in accordance with the subsidy agreement (agreement identifier 000000C313925P4H0002; grant No 139-15-2025-012).

The research was carried out using the MSU-270 supercomputer of Lomonosov Moscow State University.

We'd like to thank Valeriy Gorbachev for conducting the VLM annotation for~\Cref{sec:artifact-types}.

\bibliography{paper}
\bibliographystyle{ieeetr}

\newpage
\appendix
\input{appendix.tex}

\end{document}

%% file: appendix.tex
\section{Dataset examples}
\label{sec:dataset-examples}

\begin{figure}[t]
    \centering
    \includegraphics[width=\linewidth]{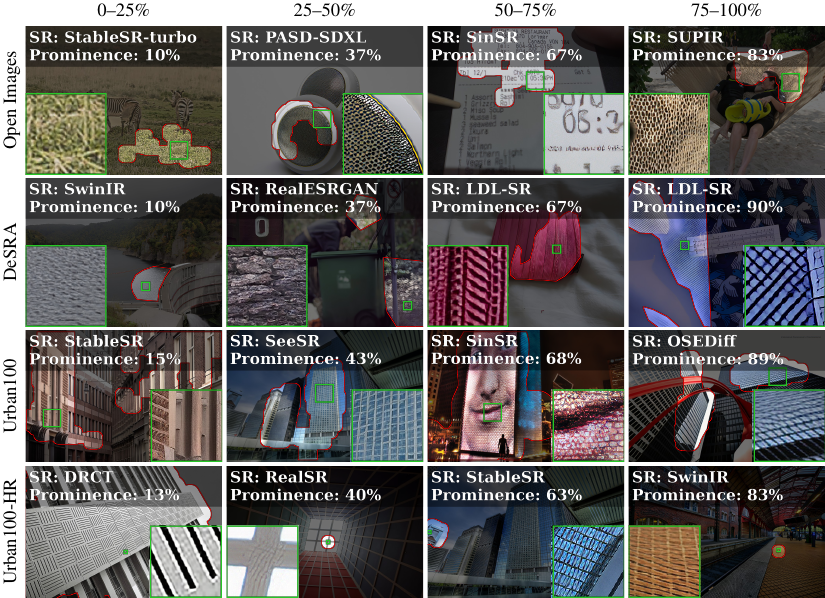}
    \caption{SR-Prominence artifact examples across components (rows) and 25\%-wide prominence ranges (columns).
    Each cell shows the full SR result with the artifact mask outlined in red and the region outside the mask dimmed.
    The green box is shown enlarged in the bottom corner.}
    \label{fig:dataset-examples}
\end{figure}

\Cref{fig:dataset-examples} shows example artifact masks from each dataset component across the prominence range.
Low-prominence examples are mostly texture distortions on water, grass, and fine repeated patterns, while higher-prominence examples include garbled text and conspicuous patterns on surfaces that should be smooth or regular.

\section{Prominence distributions}
\label{sec:prominence-distributions}

\begin{figure}[t]
    \centering
    \includegraphics[width=\linewidth]{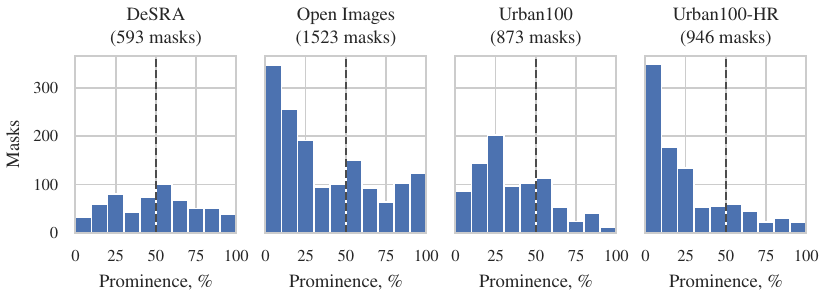}
    \caption{Distribution of crowdsourced prominence across SR-Prominence components.
    Bars show the number of masks in 10\%-wide prominence bins; the dashed line marks the 50\% cutoff for prominent masks.}
    \label{fig:prominence-histograms}
\end{figure}

\Cref{fig:prominence-histograms} shows the distribution of prominence values in each dataset component.
DeSRA is the most balanced, with in-lab artifact masks spread across the whole prominence range.
The three components with automatically collected candidate masks~(\Cref{sec:mask-collection}) are skewed toward low prominence, since many candidates proposed by the seed detectors are not very noticeable to viewers.
This skew is strongest for Urban100-HR, where the 4× larger source images make details easier to upscale~(\Cref{sec:prominence-urban100-hr}).
Nevertheless, every component contains masks across the whole prominence range, including highly prominent ones.
Despite its skew toward low prominence, Open Images contains more prominent masks than DeSRA (535 vs.\ 307), including 227 masks with prominence of at least 80\% (vs.\ 90 in DeSRA).

\section{Crowdsourced annotation dispersion analysis}
\label{sec:appendix-dispersion}

\begin{figure}[t]
    \centering
    \includegraphics[width=0.49\linewidth]{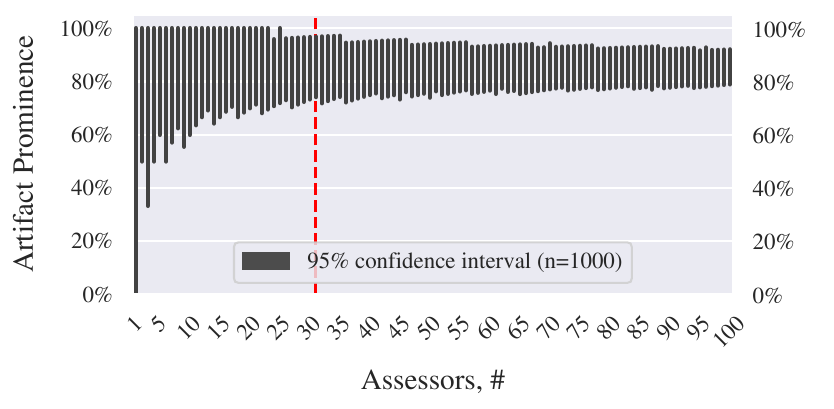}
    \includegraphics[width=0.49\linewidth]{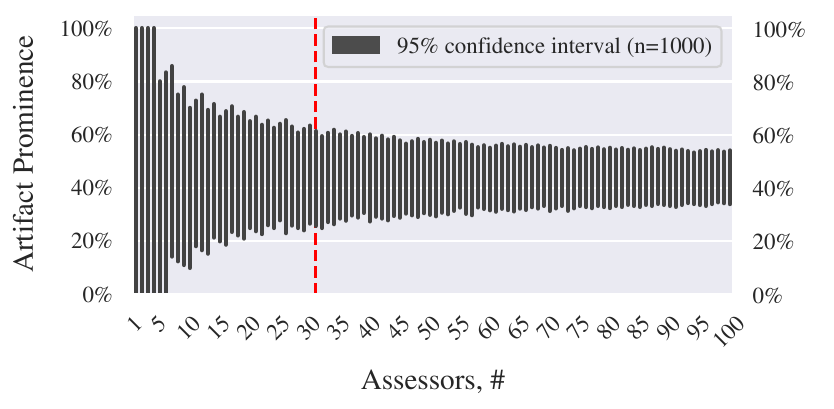}
    \caption{Bootstrap-analysis results for an image with a highly prominent artifact (left) and barely prominent artifact (right).
    Red line indicates our chosen assessor count of 30.}
    \label{fig:subj-bootstrap}
\end{figure}

We motivate our choice to use 30 assessors for every image by analyzing answer dispersion.
For this analysis, we selected 11 SR-upscaled images containing artifacts of varying intensity and conducted crowdsourced annotation following the same procedure, but with a higher participant count: every image was assessed by 250 participants instead of 30.
Next, we performed a bootstrap analysis on the votes.
For each assessor count $k$ from 1 to 100, the analysis randomly sampled $k$ votes with replacement and computed the prominence from these votes.
This procedure was repeated $n$=1000 times; we then computed 95\% confidence intervals for each assessor count $k$.

\Cref{fig:subj-bootstrap} shows these confidence intervals for two sample images: one with a highly prominent artifact and another with a barely prominent artifact.
In cases with few assessors (1--5), the confidence interval frequently spans the whole prominence range from 0\% to 100\%, meaning any given 5 assessors may all state that an artifact is present or absent.
This is especially true for unclear cases at around 50\% prominence.
By 100 assessors, the confidence interval shrinks to about \ensuremath{\pm}10\%.

For the rest of our annotation process we chose an assessor count of 30 as a reasonable compromise between the confidence of the result (\ensuremath{\pm}20\%) and the time/cost of using many assessors.

\section{Crowdsourcing worker instructions}
\label{sec:full-crowd-instructions}

The full worker instructions were as follows:
\begin{quote}
In this task, you will see images before and after upscaling.
You need to look at the upscaling result and choose whether it contains distorted objects or textures.

Pay attention to the highlighted regions; they are lighter and outlined with red boxes.

\textbf{How to complete the task:}
Look at the Original image to understand what is shown in the frame.
Then carefully examine the highlighted region in the Upscaling Result.
Choose one of the answer options.

If you see distorted objects or textures anywhere inside the highlighted region in the Upscaling Result, choose \textbf{Upscaling result has distorted objects or textures}.
If you do not see distorted objects or textures in the Upscaling Result, choose \textbf{The highlighted region does not have distorted objects or textures}.
If the image failed to load, choose \textbf{Image loading error}.

Please be careful, as the task includes control questions.

Click on the image to enlarge it and examine it more closely.
If the image is rotated, click the Rotate button.
Skip tasks where more than half of the images failed to load.
On a computer, you can use the 1, 2, 3, and arrow keys.
\end{quote}

\section{Mask-preprocessing impact on DeSRA}
\label{sec:postprocessing-desra-check}

The DeSRA dataset contains in-lab annotated masks that are not sparse and are generally suitable for human viewing as is.
We used those masks to verify the impact of our preprocessing step by running the crowdsourced prominence annotation twice: once with our preprocessing and once with unmodified masks.
For this comparison, separate groups of participants conducted the annotations, with matching question order.
The mean artifact prominence for the entire DeSRA dataset was 49.4\% with our preprocessing and 47.7\% with the original masks.
This small difference, well within annotation noise, indicates that preprocessing does not meaningfully change the outcome for masks that are already good for visual inspection.

\section{Urban100-HR cropping}
\label{sec:u100-hr-cropping}

\begin{figure}[t]
    \centering
    \includegraphics[width=\linewidth]{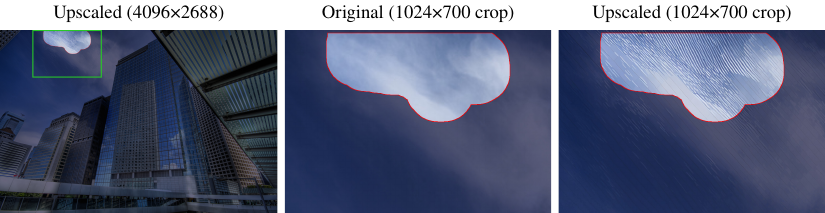}
    \caption{Urban100-HR example with a small artifact mask.
    Left: full SeeSR-turbo output with the artifact mask highlighted; the green box marks the crop.
    Middle and right: the crop of the original (nearest-neighbor upscaled input) and of the SR output, as shown to workers.
    SeeSR-turbo hallucinated diagonal streaks in the sky; the prominence of this mask is 100\%.}
    \label{fig:u100-hr-crop}
\end{figure}

For Prominence-Urban100-HR, we crop each image pair shown to workers to the bounding box of the preprocessed artifact mask, padded by 32 pixels on each side.
If the padded box is shorter than $\max(700, 0.25\,s)$ pixels along an axis, where $s$ is the image size along that axis, we extend it to that size around its center, shifting it to stay within the image.
\Cref{fig:u100-hr-crop} shows an example of a small mask on a large image: a 702×340-pixel mask on a 4096×2688 SR output, which workers saw as a 1024×700 crop.
Without cropping, the browser would display the whole image downscaled, and such small artifacts would be hard to notice.

\section{Prominent artifacts missed by seed detectors}
\label{sec:seed-detector-coverage}

\input{generated/seed_detector_coverage.tex}

Most of our artifact masks come from seed detectors~(\Cref{sec:mask-collection}), so a prominent artifact that none of the seed detectors found cannot appear in the corresponding dataset components.
However, not all of our masks come from seed detectors: all 593 masks from the DeSRA component were lab-annotated by Xie~et~al.~\citet{xie2023desra}, and also 180 masks from the Open~Images component were manually annotated by us, together comprising about 20\% of our full dataset.
We used these hand-annotated masks to estimate how many prominent artifacts our set of seed detectors could have missed.

First, we took the prominent masks from DeSRA (50\% cutoff) and ran our uncurated pipeline on their source images, producing binary masks in the same way that we used for the rest of our dataset, including mask preprocessing.
We used the same detectors listed in~\Cref{tab:summary-met-combined}, except DeSRA itself, to avoid any effect from its potential overfitting onto its own dataset.
Then, we computed how many of DeSRA's lab-annotated prominent artifact masks would be covered by our uncurated pipeline~(\Cref{tab:seed-coverage-desra}).
A mask counts as missed if at most the given fraction of its area is covered.
Only 3 of the 307 prominent DeSRA masks (1.0\%) are missed at the 50\% area coverage threshold, and none are missed at the 25\% threshold.

We also did the same experiment on the prominent masks out of the 180 hand-annotated Open~Images artifacts, which no detectors were used to find~(\Cref{tab:seed-coverage-oi}).
Here, we excluded our baseline detector since it was partially trained on them.
Only 10 of the 114 prominent masks (8.8\%) are missed at the 50\% area coverage threshold, and 3 (2.6\%) are missed at the 25\% threshold.

\begin{figure}[t]
    \centering
    \includegraphics[width=\linewidth]{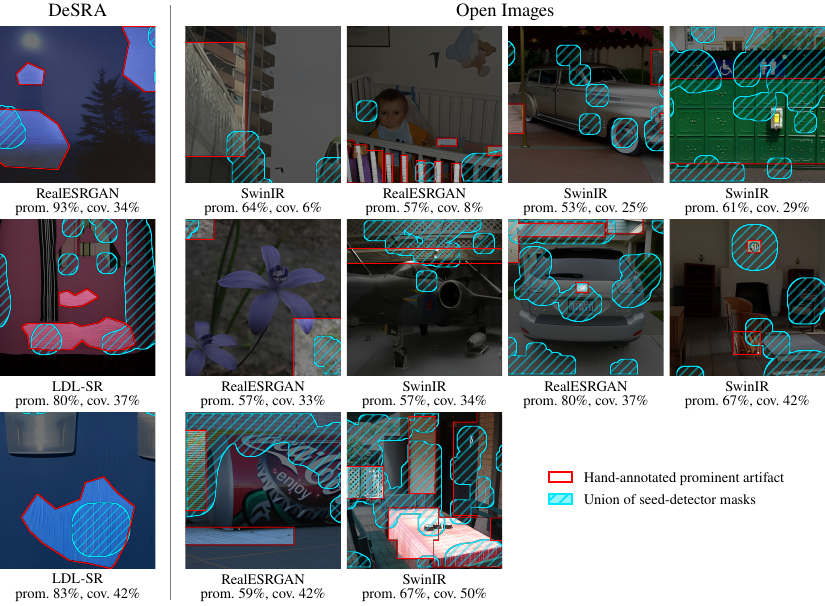}
    \caption{All prominent hand-annotated artifacts missed by the seed detectors at the 50\% area coverage threshold, sorted by coverage.
    Each cell shows a crop of the SR result around the hand-annotated mask, outlined in red, with the region outside the mask dimmed; the union of the seed-detector masks is striped in cyan.
    Labels show the SR, the crowdsourced prominence, and the fraction of the mask area covered by the seed detectors.
    Zoom in for details.}
    \label{fig:seed-missed-examples}
\end{figure}

Manual inspection of these few missed cases~(\Cref{fig:seed-missed-examples}) did not reveal a common missed artifact type.
The missed masks often occur on weakly textured surfaces, such as walls, skies, distant landscape, and interior.
These regions are also inherently difficult to delineate: the degradation changes gradually and may be confounded with natural defocus, illumination variations, or reflections.

Overall, our pipeline would find almost all of the prominent hand-annotated artifacts in both experiments.
This suggests that there is no large shared blind spot across the detectors that we used to seed our dataset.
Note however that this is not a universal guarantee across all SRs and content types.
Also, in the final dataset some masks got omitted during the top-10 strongest artifact sampling described in~\Cref{sec:mask-collection}, which allowed us to balance the crowdsourcing costs and annotate many detector-SR combinations.

\section{Prominence cutoff sensitivity}
\label{sec:cutoff-sensitivity}

\input{generated/prominence_cutoff_sensitivity.tex}

Prominence variability is one of the main challenges of accurate artifact evaluation~(\Cref{sec:appendix-dispersion}).
For this reason, we designed most of our experiments and evaluations to avoid any binary decision on the prominence: we report mean prominence in most of our tables, and our threshold-free scoring in~\Cref{tab:heatmap-prominence-contrast-reference} computes Spearman correlations directly against prominence values.

However, in \Cref{tab:summary-met-combined,tab:summary-sr-combined} we do sort detectors and SRs based on confident masks found, which uses a prominence cutoff of 50\%.
Since prominence around 50\% is uncertain, we recomputed the tables and the rankings using prominence cutoffs of 40\% and 60\%~(\Cref{tab:cutoff-detector-ranks}) to check whether this uncertainty could significantly vary these rankings.

We found that the rankings remain largely stable.
Across detectors, DISTS and SSIM retain their average rank of 2.0 at all three cutoffs and, together with the baseline, outrank the other four detectors at all cutoffs.
Across SRs, DRCT and HAT-L remain among the three SRs with the best average rank by confident masks found at all cutoffs, while LDL-SR has the worst average rank at all cutoffs (tied with RealESRGAN at 50\%).

Finally, \Cref{tab:cutoff-desra} shows how our DeSRA artifact un-noticeability headline figure changes with the prominence cutoff.

\section{Dataset scale stability}
\label{sec:dataset-scale}

\input{generated/dataset_scale_stability.tex}

To analyze how the dataset scale affects our results, we performed 1000 repetitions of random subsampling of 25\%, 50\% and 75\% of our dataset, recomputed the results and detector/SR rankings from~\Cref{tab:summary-met-combined,tab:heatmap-prominence-contrast-reference,tab:summary-sr-combined}, and computed the Spearman correlations of the resulting rankings to the rankings on our full dataset.
The subsampling was conducted by source images rather than by individual artifact masks to ensure that whole groups of similar artifacts are randomized.
\Cref{tab:scale-subsampling} shows these Spearman correlations: median and 95\% intervals across the 1000 samples.

We also performed a bootstrap over source images (random sampling with replacement until reaching our total number of source images), and the rankings remained similarly consistent.
For example, \Cref{tab:scale-bootstrap} shows the rank variations for the best and worst SRs from~\Cref{tab:summary-sr-combined}.

These results support the stability of our conclusions at our current dataset scale, while also suggesting that smaller subsets (25\%) produce substantially less reliable rankings.

\section{Artifact-type and semantic-context analysis}
\label{sec:qwen-type-context-summary}

We used the Qwen 3 VLM to assign auxiliary artifact-type labels to each annotated mask and semantic-context labels to each source image.
We then hand-checked and cleaned the artifact-type labels.
These labels are multi-valued and model-generated, so they are intended for aggregate descriptive analysis rather than as ground-truth classes.
In \Cref{tab:qwen-artifact-types,tab:qwen-semantic-contexts}, prominent rate is the fraction of masks with crowdsourced prominence $\ge 0.5$, i.e. masks for which at least half of retained workers confirmed a noticeable artifact.
Counts across rows need not sum to the dataset size because each mask or source image may receive multiple labels.

Both mean prominence and prominent rate in each cell average only over the detected artifact masks that received the given label, and are not meant to make broad statements about the data.
For example, all four face-distortion masks in Urban100-HR are prominent, but such a small sample does not show that face distortions in Urban100-HR are generally prominent.

Across the labeled datasets, artifact prominence differs substantially by artifact type.
Plastic texture is the least prominent common artifact type, appearing in 2160 masks with mean prominence 0.301 and 24.3\% prominent masks.
In contrast, hallucinated texture appears in 2377 masks with mean prominence 0.423 and 41.5\% prominent masks, suggesting that viewers are more sensitive to false structured detail, moir\'e, or invented texture than to over-smoothed or waxy surfaces.
Text aberrations are also high-prominence overall, appearing in 183 masks with mean prominence 0.473 and 48.1\% prominent masks.

Semantic context shows a similar pattern.
The highest-prominence source-image contexts are art, with 234 masks, mean prominence 0.476, and 48.7\% prominent masks, and text, with 600 masks, mean prominence 0.434 and 42.8\% prominent masks.
The lowest-prominence contexts are nature, with 536 masks, mean prominence 0.263 and 19.0\% prominent masks, and texture, with 1042 masks, mean prominence 0.293 and 22.7\% prominent masks.
Here, texture denotes images dominated by surface or material patterns, such as fabric, wood grain, asphalt, mesh, or repetitive fa\c{c}ades.

\input{generated/type_context_summary.tex}

\section{Reference baseline detection method details}
\label{sec:baseline-details}

\begin{figure*}[t]
    \centering
    \includegraphics[width=\linewidth]{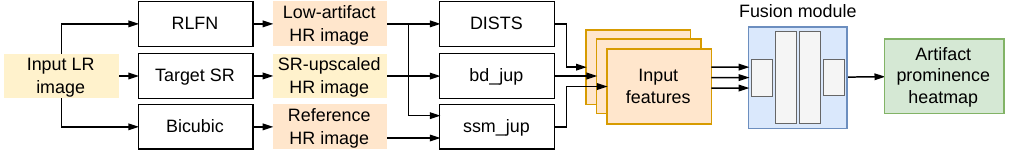}
    \caption{Architecture of the reference artifact-prominence baseline.
    The input image is upscaled by the target SR and compared against either the available HR reference or RLFN pseudo-GT as described in~\cref{sec:pseudo-gt}.
    Then, we compute three features described in~\cref{sec:feature-selection}.
    Finally, we run the fusion module described in~\cref{sec:method}.}
    \label{fig:artifact_method}
\end{figure*}

\Cref{fig:artifact_method} outlines the reference artifact-prominence baseline.
The baseline first computes three heatmaps from existing quality and artifact-detection metrics, then fuses them with a lightweight MLP into a single prominence heatmap.

\subsection{Input features}
\label{sec:feature-selection}

We selected features based on their proven performance for evaluating and detecting texture distortions.
These features estimate not only the visual quality, but also the structural similarity between the reference and the upscaled image.

The first feature is DISTS~\citep{ding2022dists}, a visual-image-quality metric which accounts for texture distortions and their perceptual impact.
As DISTS is trained on natural images, it effectively detects unnatural degradations like SR artifacts.
DISTS produces a single image-level score, so we computed it block-wise in 16×16-pixel blocks, the minimum input size of the metric.

The second feature, which we call \textit{ssm\_jup}, is adapted from the small-color-artifact detector from \citet{Tsereh2024JPEGAI}, itself based on LDL~\citep{liang2022details}.
It targets small-scale distortions and was shown to be effective for finding JPEG~AI artifacts.
To capture texture distortions, we modify the detector to use all RGB channels rather than only chromatic U and~V components.
Like LDL, this feature requires a reference image upscaled by an artifact-resistant method; we chose bicubic interpolation for this input.

The last feature, \textit{bd\_jup}, is a weighted sum of LPIPS~\citep{zhang2018unreasonable} and ERQA~\citep{kirillova2022erqa} applied block-wise.
LPIPS measures how well the upscaled image preserves perceptual quality, and is widely used in SR evaluation.
Meanwhile, ERQA assesses the preservation of object details and boundaries.
For LPIPS, we used 32×32-pixel blocks with stride of 16.
ERQA uses 8×8 blocks with no overlap.
LPIPS is weighted 3:2 compared with ERQA.

We describe the implementation of the two custom features below.

\subsection{Structure Similarity Map (ssm\_jup)}
\label{sec:ssm-jup-details}

Our \textit{ssm\_jup} feature adapts the small-color-artifact detector from~\citet{Tsereh2024JPEGAI}, which is itself based on LDL~\citep{liang2022details}.
Given a reference image $I_{\text{ref}}$ (the original HR or pseudo-GT), the SR output $I_{\text{SR}}$, and a bicubic-upscaled baseline $I_{\text{bic}}$, we compute scaled residual-variance maps for both SR and bicubic outputs.

First, we compute the absolute residual summed across channels:
\begin{equation}
R^C_x(i,j) = \sum_{c \in C} \bigl|I_x^c(i,j) - I_{\text{ref}}^c(i,j)\bigr|,
\quad x \in \{\text{SR}, \text{bic}\}.
\end{equation}

We then compute local variance within an $n \times n$ window and scale it by a global factor:
\begin{align}
M^C_x(i, j) &= var\bigl(R^C_x\left(i - \frac{n - 1}{2}:i + \frac{n - 1}{2}, j - \frac{n - 1}{2}:j + \frac{n - 1}{2}\right)\bigr),\\
S^C_x(i,j) &= var\bigl(R^C_x\bigr)^{1/5} \cdot M^C_x(i,j),
\end{align}
where $n = 33$.

The key modification from~\citet{Tsereh2024JPEGAI} is in the choice of color channels $C$: the original method computes separate maps on chrominance channels (UV from YUV and ab from Lab color spaces) and intersects the thresholded results to detect color artifacts.
We instead operate on all three RGB channels ($C = \{R, G, B\}$), which enables detection of luminance-correlated texture distortions that are common in SR artifacts but would be missed by chrominance-only analysis.

The final feature is the smoothed difference between the SR and bicubic maps:
\begin{equation}
\text{ssm\_jup} = G_\sigma * S_{\text{SR}} - G_\sigma * S_{\text{bic}},
\end{equation}
where $G_\sigma$ denotes a Gaussian kernel with $\sigma = 33$.

\subsection{Block-wise Distortion (bd\_jup)}
\label{sec:bd-jup-details}

The \textit{bd\_jup} feature combines block-wise LPIPS~\citep{zhang2018unreasonable} and ERQA~\citep{kirillova2022erqa} scores.
LPIPS is computed on $32 \times 32$ blocks with stride 16; ERQA uses $8 \times 8$ blocks with no overlap.
Since ERQA measures edge-preservation quality (higher is better), we invert it to obtain a distortion score.
The final feature is:
\begin{equation}
\text{bd\_jup} = 0.6 \cdot \text{LPIPS} + 0.4 \cdot (1 - \text{ERQA}).
\end{equation}

\subsection{Fusion module}
\label{sec:method}

We experimented with several architectures for fusing the features into a single prominence prediction, including CNN-based and tree-based models.
A shallow multilayer perceptron (MLP) achieved the best overall performance, so we adopted it as our feature fusion module.
The MLP takes as input the feature values, passes them through three fully connected layers (3-128-128-1) with ReLU activations, and outputs a single prominence value.
It independently processes each pixel of the input-feature heatmaps, yet still captures broader context since the features themselves encode both the pixel's neighborhood and wider image-level information.

\subsection{Training}

We train our fusion module using Adam on a training subset of 374 artifact examples from Prominence-OpenImages.
The model predicts a prominence value for each pixel of the input image.
We compute the mean predicted prominence inside and outside the binary artifact mask from the dataset.
The training loss consists of two $L_2$ components:
\begin{equation}
\begin{split}
\mathcal{L} &= L_2(\mathrm{MeanInside}, \mathrm{GT\ Prominence}) \\
            &+ L_2(\mathrm{MeanOutside}, 0).
\end{split}
\end{equation}
The model is trained to predict the ground-truth prominence value inside the binary mask, and 0 (no artifact) outside it.
Thanks to small model size, the training converges quickly, usually in around 10--15 epochs.
One training epoch takes about 13 seconds on an Nvidia RTX 3090 GPU.

\subsection{Training set size ablation}
\label{sec:nn-training-size}

\input{generated/nn_training_size_ablation.tex}

Our baseline is trained on only 374 artifact examples, so we studied how its performance depends on the training set size.
We trained our baseline on subsets of 25, 50, 100, and 200 masks of the original 374-mask training set (sampled by source images), as well as on the full training set, two times each with different random seeds.
Then we evaluated the resulting checkpoints using our threshold-free scoring~(\Cref{sec:score}).
\Cref{tab:nn-training-size} shows results from each training run.

These results show that the performance is very unreliable at low training set sizes (as expected), then by 200 examples it starts to stabilize, and it improves further with our original 374 examples.
Comparing with the scores of existing detectors on Urban100, an external component, in~\Cref{tab:heatmap-prominence-contrast-reference} shows that both 200-example runs outperform the specialized detectors DeSRA and LDL, suggesting that even modest prominence supervision can meaningfully improve the alignment.

\section{Subjective evaluation on additional SR datasets}
\label{sec:extra-subj-eval}

\begin{table}[t]
    \centering%
    \begin{minipage}[t]{.49\linewidth}%
\centering
\caption{Crowdsourced prominence across SR models on 6 datasets.}
\label{tab:summary_sr_other}
\resizebox{\linewidth}{!}{%
\begin{tabular}{@{}llrrr@{}}
\toprule
SR & Type & \vtop{\hbox{\strut Masks}\hbox{\strut Found}} & \vtop{\setbox0\hbox{\strut Prominence}\hbox to\wd0{\hss\strut Mean\hss}\copy0} & \vtop{\hbox{\strut Conf. Masks}\hbox{\strut Found}} \\
\midrule
SeeSR & Diffusion & 61 & \textbf{7.15\%} & 0 \\
ResShift & Diffusion & 60 & \underline{7.20\%} & 0 \\
SinSR & Diffusion & 60 & 9.19\% & 2 \\
HAT-L & Transformer & 50 & 9.60\% & 1 \\
DRCT & Transformer & 50 & 13.33\% & 0 \\
SwinIR & Transformer & 60 & 17.12\% & 4 \\
RealESRGAN & CNN & 60 & 17.46\% & 5 \\
SUPIR & Diffusion & 62 & 17.56\% & 6 \\
\bottomrule
\end{tabular}%
}%
    \end{minipage}%
    \hspace{.015\linewidth}%
    \begin{minipage}[t]{.49\linewidth}%
  \centering
  \caption{Crowdsourced prominence across artifact detection methods on 6 datasets.}
  \label{tab:summary_met_other}
\resizebox{\linewidth}{!}{%
  \begin{tabular}{@{}lrrrr@{}}
    \toprule
    Method & \vtop{\hbox{\strut Masks}\hbox{\strut Found}} & \vtop{\setbox0\hbox{\strut Prominence}\hbox to\wd0{\hss\strut Mean\hss}\copy0} & \vtop{\hbox{\strut Conf. Masks}\hbox{\strut Found}} & \vtop{\hbox{\strut Prom.~×}\hbox{\strut Conf.}} \\
    \midrule
ssm\_jup {\footnotesize (t=0.15)} & 80 & 7.42\% & 1 & 0.07 \\
bd\_jup {\footnotesize (t=0.1)} & 80 & 7.34\% & 2 & 0.15 \\
LDL {\footnotesize (t=0.005)} & 80 & 9.46\% & 2 & 0.19 \\
DeSRA & 74 & 12.21\% & 3 & 0.37 \\
\textbf{Baseline} {\footnotesize (t=0.3)} & 70 & \underline{17.85\%} & \underline{8} & \underline{1.43} \\
DISTS {\footnotesize (t=0.25)} & 76 & \textbf{18.03\%} & \textbf{9} & \textbf{1.62} \\
    \bottomrule
  \end{tabular}%
}%
    \end{minipage}%
\end{table}

We conduct an additional subjective evaluation on 6 widely known image datasets~\citep{bsds200, historical, general100, set5, set14, t91}, following the setup described in~\Cref{sec:uncurated-benchmark}.
In total, this evaluation used 420 source images, each processed by 8 SR models.

\Cref{tab:summary_sr_other,tab:summary_met_other} show the results, grouped by SR models and by artifact detection methods, respectively.
Interestingly, SR models show much better artifact robustness than in our main comparison in~\Cref{sec:experiments}, likely because these datasets are commonly used for SR training and evaluation.
Our baseline falls one confident artifact short of DISTS, but otherwise shows competitive results.

\section{Artifact examples and failure cases}

\begin{figure}[t]
    \centering
    \includegraphics[width=.9\linewidth]{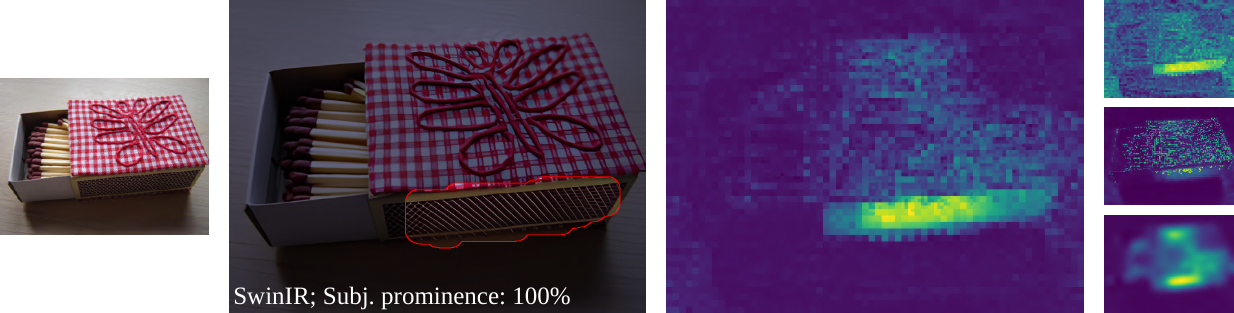}
    \includegraphics[width=.9\linewidth]{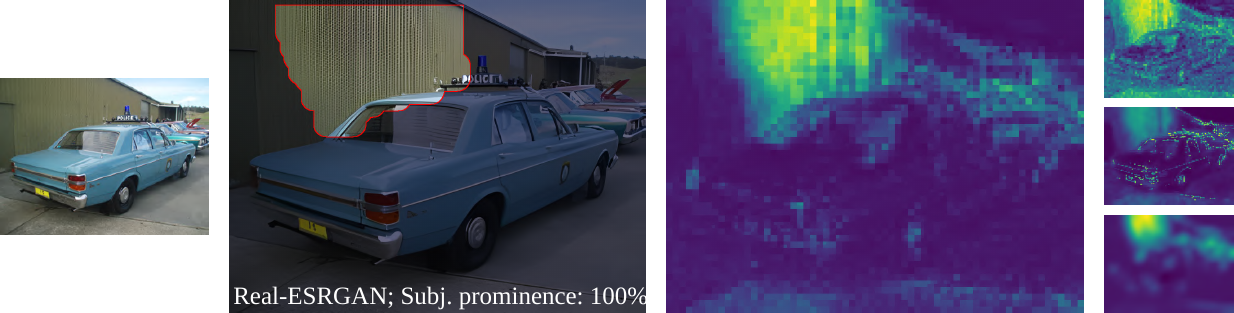}
    \includegraphics[width=.9\linewidth]{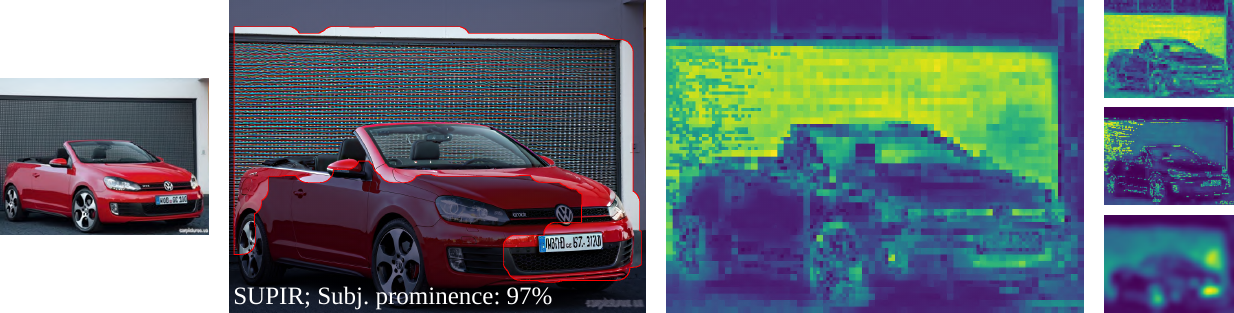}
    \includegraphics[width=.9\linewidth]{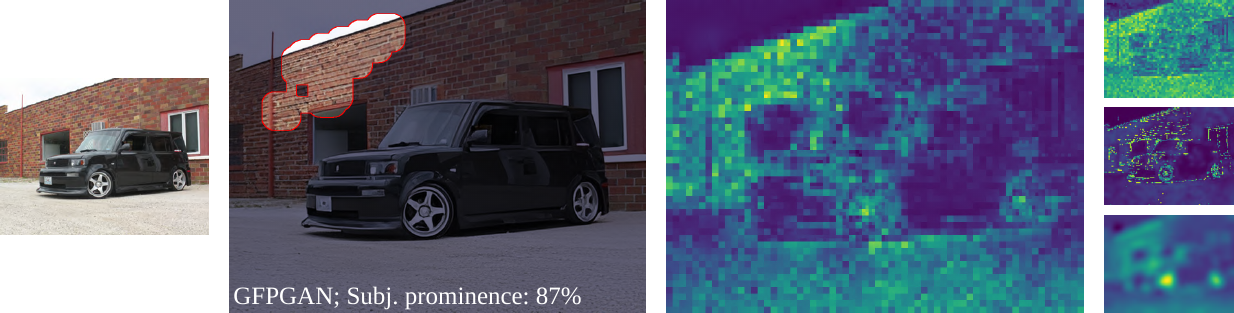}
    \includegraphics[width=.9\linewidth]{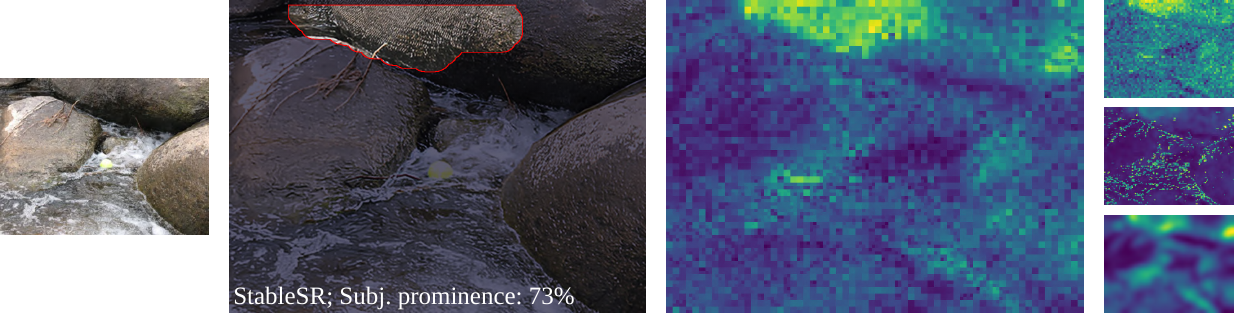}

    \vspace{5pt}
    \includegraphics[width=.9\linewidth]{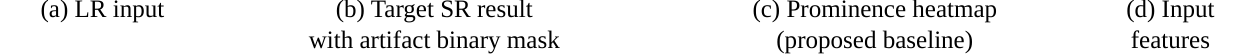}
    \caption{Example artifacts detected by the baseline.
    (a): low-resolution input image; (b):~target SR result with annotated output artifact mask; (c): artifact prominence heatmap predicted by our baseline; (d): our input features described in~Sec.~\ref{sec:feature-selection}, top to bottom: DISTS, bd\_jup, ssm\_jup.}
    \label{fig:extra-examples}
\end{figure}

\begin{figure}[p]
    \centering
    \includegraphics[width=.9\linewidth]{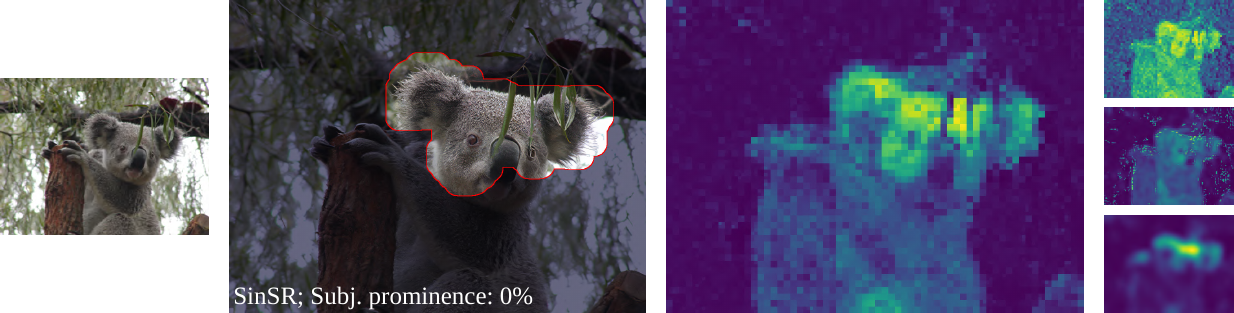}
    \includegraphics[width=.9\linewidth]{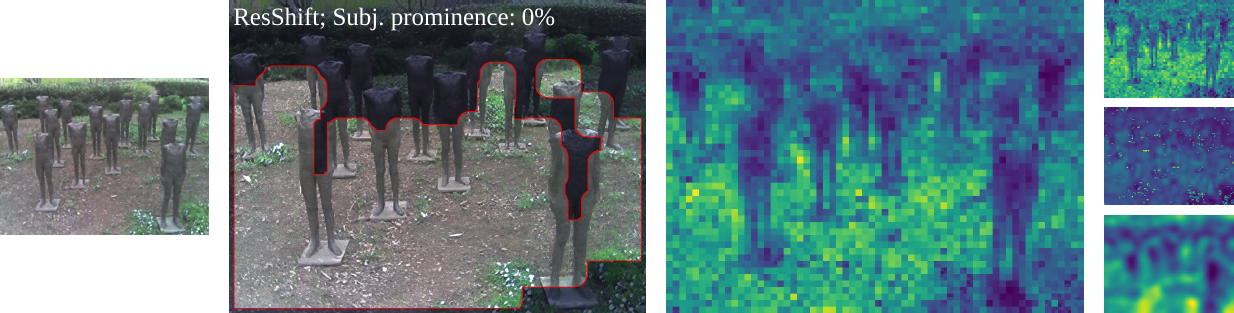}

    \includegraphics[width=.9\linewidth]{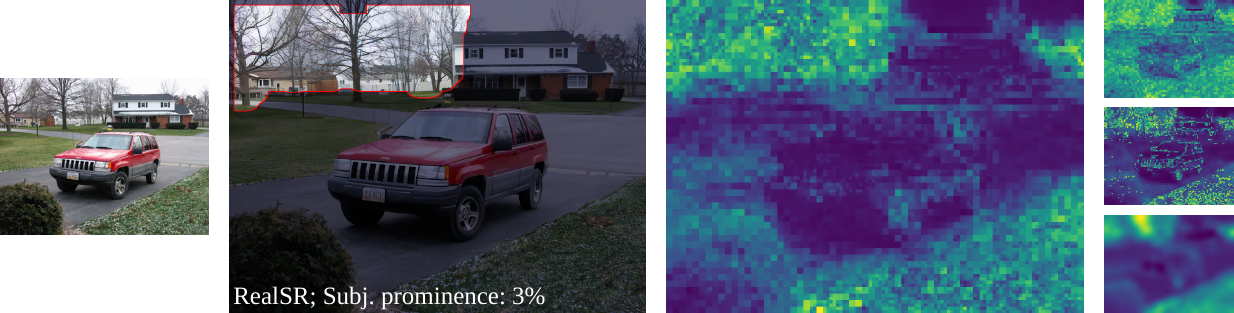}
    \includegraphics[width=.9\linewidth]{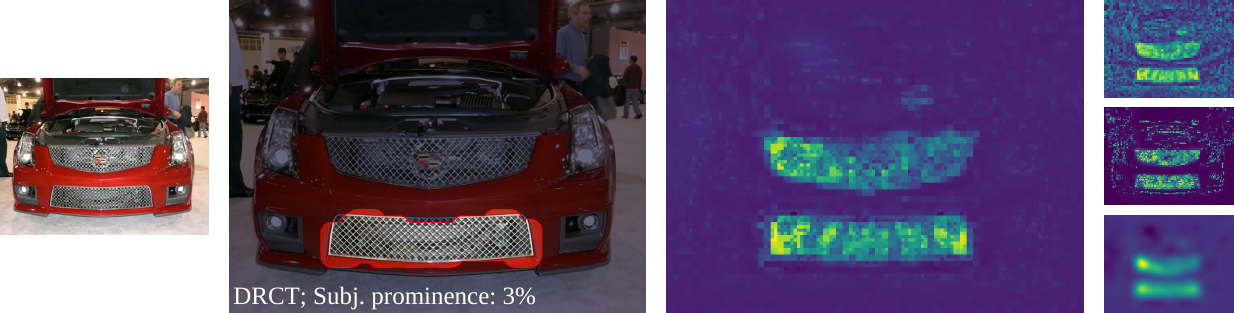}

    \vspace{5pt}
    \includegraphics[width=.9\linewidth]{img/ex/labels.pdf}
    \caption{Example false detections by the baseline.
    (a): low-resolution input image; (b):~target SR result with annotated output artifact mask; (c): artifact prominence heatmap predicted by our baseline; (d): our input features described in~Sec.~\ref{sec:feature-selection}, top to bottom: DISTS, bd\_jup, ssm\_jup.}
    \label{fig:failure-cases}
\end{figure}

\begin{figure}[t]
    \centering
    \includegraphics[width=\linewidth]{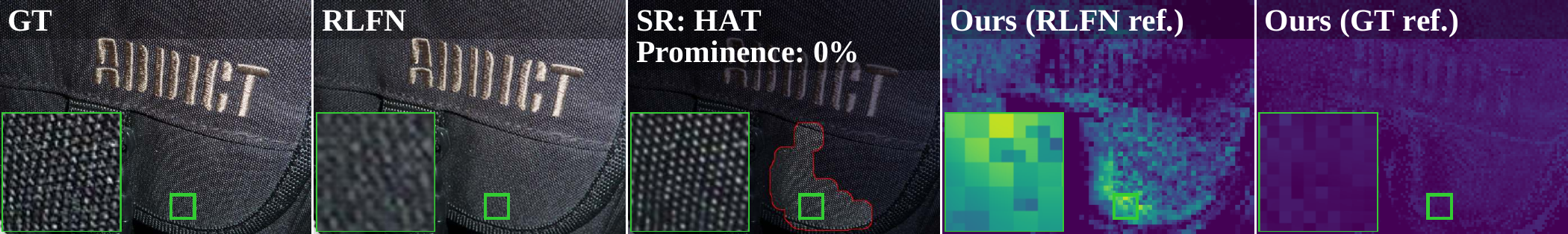}
    \includegraphics[width=\linewidth]{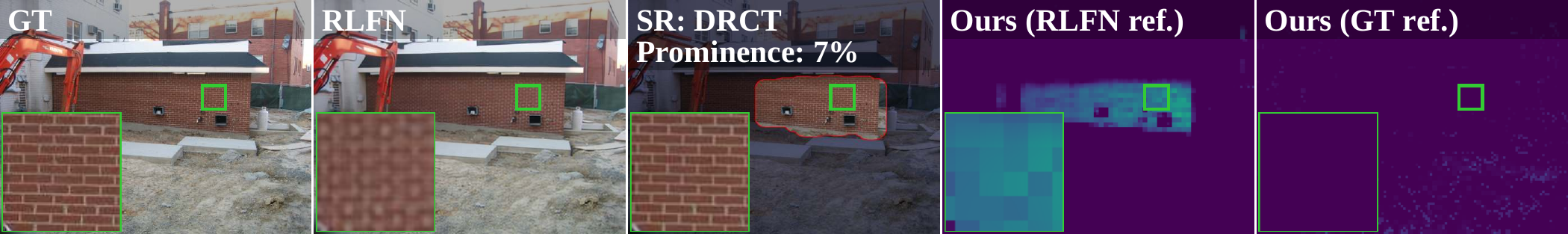}
    \caption{Example false detections by the baseline due to inaccurate restoration from pseudo-GT lightweight SR (RLFN).
    Rightmost column shows that the false detection disappears when an accurate restoration is used as reference instead of RLFN.}
    \label{fig:fp-cases-rlfn}
\end{figure}

\Cref{fig:extra-examples} shows examples of prominent artifacts detected by our baseline across various SR models~\citep{yu2024scaling,wang2021real,wang2024exploiting,liang2021swinir,wang2021towards}.
Each example is annotated with the binary artifact mask and subjective prominence.

\Cref{fig:failure-cases} shows examples of false detections by our baseline across SR models~\citep{Hsu_2024_CVPR,yue2023resshift,wang2024sinsr,Ji_2020_RealSR}.
We observed the following failure cases:
\begin{itemize}
    \item Distortions on natural, unstructured objects, like ground, grass, or trees, that are not very prominent to human observers.
    \item Accurate restoration of fine textures such as fur, nylon, or mesh grille.
    False detections can happen on these when the lightweight SR (in our case, RLFN~\citep{Fangyuan2022RLFN}) fails to produce a sharp upscaling of the texture, leading the metrics to see a discrepancy to the target SR and mark it as an artifact.
    Using an accurately-restored reference removes those false detections as~\Cref{fig:fp-cases-rlfn} shows.
\end{itemize}

Existing methods also suffer from these failure cases; indeed, they account for most of the low-prominence detections from our subjective evaluation described in~\Cref{sec:findings-metrics}.

%% file: generated/seed_detector_coverage.tex
\begin{table}[t]
\centering
\begin{minipage}[t]{.48\linewidth}
\centering
\caption{Prominent DeSRA in-lab artifact masks missed by the seed detectors.}
\label{tab:seed-coverage-desra}
\begin{tabular}{@{}lrr@{}}
\toprule
\textbf{Coverage threshold} & \textbf{Missed} & \textbf{Missed \%} \\
\midrule
25\% area covered & 0 & 0.0\% \\
50\% area covered & 3 & 1.0\% \\
\bottomrule
\end{tabular}
\end{minipage}
\hfill
\begin{minipage}[t]{.48\linewidth}
\centering
\caption{Prominent hand-annotated Open~Images artifact masks missed by the seed detectors.}
\label{tab:seed-coverage-oi}
\begin{tabular}{@{}lrr@{}}
\toprule
\textbf{Coverage threshold} & \textbf{Missed} & \textbf{Missed \%} \\
\midrule
25\% area covered & 3 & 2.6\% \\
50\% area covered & 10 & 8.8\% \\
\bottomrule
\end{tabular}
\end{minipage}
\end{table}

%% file: generated/prominence_cutoff_sensitivity.tex
\begin{table}[t]
\centering
\begin{minipage}[t]{.52\linewidth}
\centering
\caption{Average rank by Prom.~$\times$~Conf.$^\uparrow$ across detection methods (\Cref{tab:summary-met-combined}) at different prominence cutoffs.}
\label{tab:cutoff-detector-ranks}
\begin{tabular}{@{}lrrr@{}}
\toprule
\textbf{Method} & \textbf{40\%} & \textbf{50\%} & \textbf{60\%} \\
\midrule
bd\_jup {\footnotesize (t=0.1)} & 5.7 & 6.3 & 6.3 \\
LDL {\footnotesize (t=0.005)} & 5.7 & 5.7 & 5.3 \\
ssm\_jup {\footnotesize (t=0.15)} & 6.0 & 5.3 & 5.3 \\
DeSRA & 4.7 & 4.3 & 4.7 \\
\textbf{Baseline} {\footnotesize (t=0.3)} & \textbf{2.0} & 2.3 & 2.3 \\
DISTS {\footnotesize (t=0.25)} & \textbf{2.0} & \textbf{2.0} & \textbf{2.0} \\
SSIM {\footnotesize (t=0.55)} & \textbf{2.0} & \textbf{2.0} & \textbf{2.0} \\
\bottomrule
\end{tabular}
\end{minipage}%
\hspace{.04\linewidth}%
\begin{minipage}[t]{.44\linewidth}
\centering
\caption{Share of DeSRA in-lab artifact masks noticed by fewer viewers than each prominence cutoff.}
\label{tab:cutoff-desra}
\begin{tabular}{@{}lr@{}}
\toprule
\textbf{Cutoff} & \textbf{Masks below cutoff} \\
\midrule
40\% & 213 / 593 (35.9\%) \\
50\% & 286 / 593 (48.2\%) \\
60\% & 367 / 593 (61.9\%) \\
\bottomrule
\end{tabular}
\end{minipage}
\end{table}

%% file: generated/dataset_scale_stability.tex
\begin{table*}[t]
\centering
\caption{Spearman correlations of average ranks on random source-image subsets to those on the full dataset, median [95\% interval] across 1000 samples.}
\label{tab:scale-subsampling}
\resizebox{\textwidth}{!}{%
\begin{tabular}{@{}lcccc@{}}
\toprule
\textbf{Source images retained} & \textbf{Detector Prom.~$\times$~Conf.} & \textbf{SR Conf. Masks} & \textbf{SR Mean Prominence} & \textbf{Threshold-free score} \\
\midrule
25\% & 0.87 [0.47, 0.99] & 0.84 [0.66, 0.93] & 0.84 [0.69, 0.95] & 0.91 [0.70, 0.98] \\
50\% & 0.93 [0.71, 1.00] & 0.93 [0.83, 0.97] & 0.91 [0.82, 0.98] & 0.96 [0.85, 0.99] \\
75\% & 0.97 [0.81, 1.00] & 0.96 [0.92, 0.99] & 0.96 [0.90, 0.99] & 0.98 [0.92, 1.00] \\
\bottomrule
\end{tabular}%
}
\end{table*}

\begin{table}[t]
\centering
\caption{Average mean-prominence rank of the best and worst SR models under source-image bootstrap, median [95\% interval] across 1000 samples.}
\label{tab:scale-bootstrap}
\begin{tabular}{@{}lc@{}}
\toprule
\textbf{SR} & \textbf{Average mean-prominence rank} \\
\midrule
DRCT & 1.33 [1.00, 3.00] \\
HAT-L & 2.33 [1.33, 4.33] \\
LDL-SR & 16.33 [15.00, 16.67] \\
\bottomrule
\end{tabular}
\end{table}

%% file: generated/type_context_summary.tex
\begin{table*}[t]
\centering
\caption{Qwen 3 VLM artifact-type labels across SR-Prominence. Each cell reports $n$ / mean prominence / prominent rate.}
\label{tab:qwen-artifact-types}
\resizebox{\textwidth}{!}{%
\begin{tabular}{@{}lrrrrr@{}}
\toprule
\textbf{Artifact type} & \textbf{DeSRA} & \textbf{OpenImages} & \textbf{Urban100} & \textbf{Urban100-HR} & \textbf{Total} \\
\midrule
hallucinated texture & 325 / 0.580 / 68.0\% & 812 / 0.479 / 50.0\% & 701 / 0.398 / 34.0\% & 539 / 0.274 / 22.4\% & 2377 / 0.423 / 41.5\% \\
plastic texture & 364 / 0.431 / 42.3\% & 934 / 0.283 / 22.7\% & 416 / 0.323 / 23.8\% & 446 / 0.212 / 13.5\% & 2160 / 0.301 / 24.3\% \\
text aberrations & -- & 146 / 0.523 / 56.8\% & 27 / 0.281 / 7.4\% & 10 / 0.263 / 30.0\% & 183 / 0.473 / 48.1\% \\
face distortion & -- & 42 / 0.455 / 42.9\% & 16 / 0.424 / 25.0\% & 4 / 0.983 / 100.0\% & 62 / 0.481 / 41.9\% \\
color change & 4 / 0.425 / 50.0\% & 8 / 0.263 / 12.5\% & 10 / 0.336 / 30.0\% & 2 / 0.383 / 0.0\% & 24 / 0.330 / 25.0\% \\
deblurring & 14 / 0.233 / 14.3\% & -- & -- & -- & 14 / 0.233 / 14.3\% \\
other & -- & -- & -- & 3 / 0.078 / 0.0\% & 3 / 0.078 / 0.0\% \\
\bottomrule
\end{tabular}%
}
\end{table*}

\begin{table*}[t]
\centering
\caption{Qwen 3 VLM semantic-context labels across SR-Prominence. Each cell reports $n$ / mean prominence / prominent rate.}
\label{tab:qwen-semantic-contexts}
\resizebox{\textwidth}{!}{%
\begin{tabular}{@{}lrrrrr@{}}
\toprule
\textbf{Semantic context} & \textbf{DeSRA} & \textbf{OpenImages} & \textbf{Urban100} & \textbf{Urban100-HR} & \textbf{Total} \\
\midrule
urban & 91 / 0.379 / 35.2\% & 455 / 0.385 / 36.3\% & 822 / 0.366 / 29.0\% & 885 / 0.245 / 18.3\% & 2253 / 0.323 / 26.5\% \\
objects & 390 / 0.512 / 55.1\% & 1043 / 0.426 / 42.8\% & 205 / 0.368 / 29.3\% & 425 / 0.222 / 15.3\% & 2063 / 0.394 / 38.1\% \\
texture & 94 / 0.538 / 58.5\% & 259 / 0.328 / 28.2\% & 322 / 0.313 / 20.5\% & 367 / 0.188 / 11.7\% & 1042 / 0.293 / 22.7\% \\
people & 120 / 0.483 / 48.3\% & 257 / 0.383 / 38.9\% & 197 / 0.368 / 26.9\% & 191 / 0.307 / 29.8\% & 765 / 0.376 / 35.0\% \\
text & 79 / 0.521 / 57.0\% & 343 / 0.458 / 48.1\% & 110 / 0.350 / 24.5\% & 68 / 0.346 / 29.4\% & 600 / 0.434 / 42.8\% \\
nature & 131 / 0.433 / 39.7\% & 370 / 0.211 / 13.2\% & 24 / 0.205 / 0.0\% & 11 / 0.121 / 9.1\% & 536 / 0.263 / 19.0\% \\
animals & 219 / 0.487 / 52.1\% & 197 / 0.277 / 20.3\% & -- & -- & 416 / 0.387 / 37.0\% \\
action & 68 / 0.474 / 50.0\% & 107 / 0.313 / 26.2\% & 39 / 0.410 / 38.5\% & 90 / 0.330 / 36.7\% & 304 / 0.366 / 36.2\% \\
art & 12 / 0.527 / 75.0\% & 99 / 0.400 / 39.4\% & 101 / 0.589 / 63.4\% & 22 / 0.270 / 9.1\% & 234 / 0.476 / 48.7\% \\
\bottomrule
\end{tabular}%
}
\end{table*}

%% file: generated/nn_training_size_ablation.tex
\begin{table*}[t]
\centering
\caption{Threshold-free prominence score (\Cref{sec:score}) of the baseline trained on random subsets of its training set. Each cell reports two runs with different random training subsets and initializations; the mean is over both runs and all columns.}
\label{tab:nn-training-size}
\resizebox{\textwidth}{!}{%
\begin{tabular}{@{}lrrrrrrr@{}}
\toprule
 & \multicolumn{2}{c}{\textbf{Open Images}} & \multicolumn{2}{c}{\textbf{Urban100}} & \multicolumn{2}{c}{\textbf{DeSRA}} & \\ \cmidrule(lr){2-3} \cmidrule(lr){4-5} \cmidrule(lr){6-7}
\textbf{Training examples} & \textbf{Original HR} & \textbf{RLFN} & \textbf{Original HR} & \textbf{RLFN} & \textbf{MSE-SR} & \textbf{RLFN} & \textbf{Mean} \\
\midrule
25 & 0.139 / 0.408 & 0.068 / 0.277 & 0.066 / 0.192 & 0.133 / 0.182 & -0.245 / 0.210 & -0.213 / 0.217 & 0.120 \\
50 & 0.149 / 0.385 & 0.084 / 0.243 & 0.157 / 0.167 & 0.162 / 0.152 & -0.153 / 0.194 & -0.140 / 0.216 & 0.135 \\
100 & 0.414 / 0.388 & 0.276 / 0.246 & 0.212 / 0.199 & 0.182 / 0.178 & 0.231 / 0.107 & 0.241 / 0.129 & 0.233 \\
200 & 0.453 / 0.475 & 0.278 / 0.364 & 0.249 / 0.295 & 0.216 / 0.239 & 0.184 / 0.178 & 0.181 / 0.187 & 0.275 \\
374 & 0.500 / 0.517 & 0.322 / 0.362 & 0.288 / 0.297 & 0.255 / 0.247 & 0.143 / 0.158 & 0.134 / 0.169 & 0.283 \\
\bottomrule
\end{tabular}%
}
\end{table*}